\documentclass{IEEEoj}

\usepackage{cite}
\usepackage{amsmath,amssymb,amsfonts}
\usepackage{algorithmic}
\usepackage{graphicx}
\usepackage{textcomp}
\usepackage{tabularx}
\usepackage{multirow}
\usepackage[hidelinks]{hyperref}
\usepackage{epstopdf}
\usepackage{subcaption}
\usepackage{array}
\usepackage{booktabs}
\usepackage{stfloats}
\usepackage{url}
\usepackage{verbatim}
\usepackage{makecell}
\usepackage{threeparttable}
\usepackage{float}
\usepackage{placeins}
\usepackage[acronym,nomain,nonumberlist]{glossaries}
\glsdisablehyper
\usepackage{bm}
\usepackage{xcolor}
\usepackage{colortbl}
\AtBeginDocument{\definecolor{ojcolor}{cmyk}{0.93,0.59,0.15,0.02}}
\def\OJlogo{}
\def\BibTeX{{\rm B\kern-.05em{\sc i\kern-.025em b}\kern-.08em
    T\kern-.1667em\lower.7ex\hbox{E}\kern-.125emX}}
\makeatletter
\def\titlefont{\fontfamily{\sfdefault}\fontsize{22}{24}\selectfont\bfseries\leftskip1pc plus1fill\rightskip1pc plus1fill}
\makeatother

\newacronym{MHA}{MHA}{multi-head attention mechanism}
\newacronym{MSE}{MSE}{mean square error}
\newacronym{NMSE}{NMSE}{normalized mean square error}
\newacronym{MAE}{MAE}{mean absolute error}
\newacronym{RCM}{RCM}{reflection coefficient magnitude}
\newacronym{WI}{WI}{Wireless Insite}
\newacronym{RT}{RT}{ray tracing}
\newacronym{CIR}{CIRs}{channel impulse responses}
\newacronym{GPU}{GPU}{graphics processing unit}
\newacronym{SCM}{SNGSMCM}{statistical channel modeling}
\newacronym{EM}{EM}{electromagnetic}
\newacronym{ResNet}{ResNet}{residual neural network}
\newacronym{NGSM}{NGSM}{non-geometry stochastic modeling}
\newacronym{GBSM}{GBSM}{geometry-based stochastic modeling}
\newacronym{GBDM}{GBDM}{geometry-based deterministic modeling}
\newacronym{AI}{AI}{artificial intelligence}

\definecolor{myred}{HTML}{C00000}

\renewcommand{\bottomfraction}{0.85}
\renewcommand{\textfraction}{0.07}

\newcommand{\tightdbltopfloat}{%
    \setlength{\textfloatsep}{0.25\baselineskip}%
    \setlength{\dbltextfloatsep}{0.25\baselineskip}%
    \setlength{\floatsep}{0.25\baselineskip}%
    \setlength{\dblfloatsep}{0.25\baselineskip}%
    \setlength{\intextsep}{0.25\baselineskip}%
}

\begin{document}

\title{GeNeRT: A Physics-Informed Approach to Intelligent Wireless Channel Modeling via Generalizable Neural Ray Tracing}

\author{Kejia Bian\affilmark{1,2}, Meixia Tao\affilmark{1}, Shu Sun\affilmark{1}, Tongjia Zhang\affilmark{1}, and Jun Yu\affilmark{1}}
\affil{School of Information Science and Electronic Engineering, Shanghai Jiao Tong University, Shanghai 200240, China}
\affil{Shanghai Innovation Institute, Shanghai 200231, China}
\authornote{A preliminary version of this work was accepted at the IEEE/CIC ICCC, Shanghai, China, Aug. 2025 \cite{conferenceWork}.}
\markboth{GeNeRT: Generalizable Neural Ray Tracing}{Bian \textit{et al.}}
\begin{abstract}

Neural ray tracing (RT) has emerged as a promising paradigm for channel modeling by integrating physical propagation principles with neural networks. However, existing neural RT methods remain limited by strong spatial dependence and weak adherence to electromagnetic laws. We propose GeNeRT, a generalizable neural RT framework that improves generalization and accuracy through relative geometric features, scatterer semantics, and a Fresnel-inspired polarization-driven architecture. GeNeRT is trained through a three-stage strategy: polarization-specific module-wise pre-training captures general ray-surface interaction behavior; system-wise end-to-end training uses only receiver-side channel impulse responses to learn site-specific propagation characteristics; and measurement-based fine-tuning employs sparse measured multipath components (MPCs) to adapt polarization-related modules to real-world environments. Extensive outdoor simulations demonstrate robust intra-scenario transferability and inter-scenario zero-shot generalization. In an unseen scenario, GeNeRT achieves an overall error of $-35.36$ dB and an average-delay error of 4.91 ns, compared with $-10.85$ dB and 32.38 ns for the best baseline. With only 75 measured reflected MPCs, fine-tuning further reduces the overall error from $-14.48$ to $-22.90$ dB and the average-delay error from 6.28 to 3.58 ns. Ablation studies confirm the effectiveness of the proposed architecture and training strategy.

\end{abstract}


\maketitle

\section{Introduction}
\label{sec:introduction}
\IEEEPARstart{W}{ireless} channel modeling serves as a fundamental prerequisite for modern communication system design. By accurately characterizing wireless radio propagation mechanisms, such as path loss, multipath fading, and delay spread, channel models drive critical processes ranging from architectural design to performance evaluation and network planning. As the sixth-generation (6G) wireless systems advance into new frontiers encompassing new spectrum bands, space-air-ground-sea integrated networks, low-altitude economy applications, and integrated communication and sensing functionalities, they demand novel channel modeling methods that simultaneously achieve high accuracy, low complexity, and strong generalization.

Conventional channel modeling methods can be divided into three major categories: stochastic channel modeling (SCM), deterministic channel modeling (DCM), and hybrid channel modeling (HCM). SCM, including non-geometry and geometry based stochastic modeling, abstracts the random characteristics of the channel by probabilistic models \cite{wang2018survey}. This modeling approach has been widely used in standardized channel models, such as 3GPP TR 38.901 \cite{3gpp38901}. However, SCM can only provide empirical distributions of channel parameters for general scenarios and cannot provide site-specific channel characteristics. In contrast, DCM leverages deterministic methods to achieve higher accuracy and site-specific realism through map-based modeling. Typical deterministic methods are full-wave simulation \cite{10633853} and \gls{RT} \cite{he2018design}. Among them, RT predicts radio wave propagation by simulating physical interactions of rays with objects in three-dimensional (3D) environments. RT is much more computationally efficient and hence widely adopted in practical wireless system design. Nevertheless, RT suffers from potentially inaccurate approximations of Maxwell’s equations, particularly in complex or irregular environments. HCM combines the strengths of SCM and DCM to achieve a trade-off between accuracy and computational complexity. One popular type of HCM uses \gls{RT} to construct dominant propagation paths, while employing different stochastic approaches to model the remaining components as in \cite{7508965, 7063558}.  However, the persistent dependence on statistical elements remains a constraint on the accuracy of HCM.

Recent advances in \gls{AI} have brought increasing attention to intelligent channel modeling (ICM) \cite{9713743, 10599118, 10930391, li2025deeprt, 10949588}, which leverages machine learning techniques to learn environment features, thereby achieving more accurate and site-specific channel modeling. This line of work can be broadly classified into three categories. The first employs neural networks to directly map environmental features to channel characteristics in an end-to-end manner. For instance, in \cite{9354041}, a UNet-based convolution neural network (CNN) is employed to model and predict path loss distributions in urban environments by taking a 3D city map as input. To improve prediction accuracy in urban street canyons, the authors in \cite{9722715} augment 3D building mesh data with LiDAR point cloud-based street clutter information, and adopt a CNN-based autoencoder to extract relevant spatial features. However, the aforementioned methods \cite{9354041, 9722715} often suffer from a lack of interpretability and limited generalization capabilities.

The second category leverages wireless radiation field reconstruction techniques to predict the 3D spatial distribution of signal strength. Specifically, $\text{NeRF}^2$ in \cite{zhao2023nerf2} introduces a voxel-based radiance field representation, where dual multilayer perceptrons (MLPs) predict the attenuation and emission characteristics at each voxel. Note that $\text{NeRF}^2$ suffers from limited scalability due to dense real-world measurements. To alleviate this issue, NeWRF in \cite{lu2024newrf} proposes a direction-aware training strategy that incorporates angle-of-arrival (AoA) priors into the learning process. This design not only reduces data requirements but also improves training efficiency. Building upon this direction, WRF-GS \cite{WRF_GS} further enhances computational efficiency and mitigates rendering latency by integrating 3D Gaussian splatting with neural modeling in the radio-frequency (RF) domain. While these methods enable efficient and accurate channel modeling, they still require substantial training data and exhibit limited generalization under environmental changes.

The third category of ICM is neural \gls{RT}, which integrates neural networks into the \gls{RT} framework to implicitly model complex radio propagation within real-world environments and their geometric structures. By design, neural \gls{RT}-based approaches can offer better generalization than the first two categories. They also have the potential to achieve higher modeling accuracy than conventional \gls{RT} by learning from measured channel data rather than relying solely on simplified \gls{EM} assumptions. An early work on neural \gls{RT} is WiNeRT \cite{WiNeRT}, which employs MLPs to jointly predict ray-surface interactions and ray trajectories along propagation paths. Specifically, at each interaction, the neural network maps the incident ray and surface properties to the outgoing ray state, including its propagation direction and attenuation. Building on this line of ray trajectory modeling, SANDWICH \cite{jin2024sandwich} formulates ray tracing as a sequential decision-making problem and adopts a decision Transformer-based approach to improve both the prediction accuracy and computational efficiency of ray trajectory generation. In a similar vein, \cite{Eertmans2026TransformInvariant} introduces transform-invariant generative ray-path sampling, further reducing the substantial computational cost associated with tracing high-order ray trajectories. From the perspective of ray-surface interaction modeling, the work in \cite{LWDT} introduces learnable wireless digital twins (LWDT), which provides a more comprehensive and polarization-aware neural representation of ray-surface interactions under different propagation mechanisms. In LWDT, each propagation mechanism is modeled by a specialized neural module, while each interactive object is encoded by a dedicated neural network. Compared with WiNeRT \cite{WiNeRT}, LWDT therefore enables more fine-grained modeling of mechanism-specific and object-specific \gls{EM} interaction behavior.

Despite their advancements in modeling accuracy and efficiency, most existing neural RT methods \cite{WiNeRT, jin2024sandwich, Eertmans2026TransformInvariant, LWDT} still exhibit limited and inconsistent generalization when applied to unseen receiver (Rx) locations or new scenarios. 
SANDWICH demonstrates reasonable generalization for unseen transmitter (Tx) and Rx placements, but suffers from environmental dependence due to its autoregressive and reinforcement learning design.
In a related direction, the work in \cite{Eertmans2026TransformInvariant} accelerates point-to-point RT through learned ray-path sampling, but tests on real-world urban geometries indicate limited transfer from procedurally generated street-canyon training scenes.
WiNeRT generalizes well to diverse indoor layouts, yet its reliance on absolute spatial information, such as normal vectors and input directions, results in significant spatial dependence.
LWDT employs relative geometric features and object-level modeling, which improve its adaptability to unfamiliar indoor settings. 
However, when extended to outdoor environments, LWDT is constrained by sparse training data and tends to learn only local features of objects, which hinders comprehensive modeling of the overall \gls{EM} behavior.

\begin{table*}[!t]
    \caption{Summary of main notations.}
    \centering
    \footnotesize
    \renewcommand{\arraystretch}{1.03}
    \setlength{\tabcolsep}{3pt}
    \begin{tabularx}{0.98\textwidth}{
        >{\centering\arraybackslash}m{2.55cm}
        >{\arraybackslash}X}
        \toprule
        \textbf{Notation} & \textbf{Description} \\
        \midrule
        $i$, $j$ & Integer indices of Tx and Rx, respectively. \\
        $n$, $m$, $k$ & Integer indices of propagation path, ray, and ray--surface interaction, respectively. \\
        $N_t$, $N_r$ & Integer scalars denoting the numbers of Txs and Rxs, respectively. \\
        $N_{i,j}$ & Integer scalar denoting the number of propagation paths between the $i$-th Tx and the $j$-th Rx. \\
        $\hat{N}_{i,j}$ & Integer scalar denoting the predicted number of propagation paths between the $i$-th Tx and the $j$-th Rx. \\
        $M$ & Integer scalar denoting the number of rays initialized from the Tx. \\
        $I_n$, $I_m$ & Integer scalars denoting the number of ray--surface interactions along the $n$-th path and the number of interactions completed by the $m$-th ray, respectively. \\
        \midrule
        $h(\tau,\mathbf{\Theta},\mathbf{\Phi})$ & Complex-valued omnidirectional CIR indexed by delay, AoD, and AoA. \\
        $\mathcal{H}_n$ & Tuple describing the $n$-th MPC, i.e., $\{\tau_n,\mathbf{\Theta}_n,\mathbf{\Phi}_n,a_n\}$. \\
        $a_n$ & Complex scalar denoting the total attenuation coefficient of the $n$-th propagation path. \\
        $\tau_n$ & Real scalar denoting the delay of the $n$-th propagation path. \\
        $\mathbf{\Theta}_n$ & Real vector denoting the azimuth/elevation AoD of the $n$-th propagation path. \\
        $\mathbf{\Phi}_n$ & Real vector denoting the azimuth/elevation AoA of the $n$-th propagation path. \\
        $d_n$ & Real scalar denoting the total propagation distance of the $n$-th path. \\
        $\Gamma_n^{(k)}$ & Complex scalar denoting the interaction attenuation coefficient at the $k$-th interaction of the $n$-th path. \\
        $\mathcal{P}_n^{(I_n)}$ & Tuple of real 3D vectors denoting the ordered interaction-point coordinates along the $n$-th path. \\
        $\mathbf{p}_n^{(k)}$ & Real 3D vector denoting the coordinate of the $k$-th interaction point on the $n$-th path. \\
        \midrule
        $\mathcal{R}_m^{(k)}$ & Tuple denoting the state of the $m$-th ray after the $k$-th interaction. \\
        $\mathcal{P}_m^{(k)}$ & Tuple of real 3D vectors denoting the ordered interaction-point coordinates of the $m$-th ray up to the $k$-th interaction. \\
        $\mathbf{p}_m^{(k)}$ & Real 3D vector denoting the coordinate of the $k$-th interaction point of the $m$-th ray. \\
        $\mathbf{d}_m^{(k)}$ & Real 3D unit vector denoting the propagation direction of the $m$-th ray after the $k$-th interaction. \\
        $\mathbf{v}_m^{(k)}$ & Real 3D unit vector denoting the surface normal vector at the $k$-th interaction of the $m$-th ray. \\
        $\Gamma_m^{(k)}$ & Complex scalar denoting the interaction attenuation coefficient at the $k$-th interaction of the $m$-th ray. \\
        $\prod_{l=1}^{k}\Gamma_m^{(l)}$ & Complex scalar denoting the cumulative interaction coefficient of the $m$-th ray up to the $k$-th interaction. \\
        $\alpha_m^{(k)}$ & Real scalar denoting the incident angle of the $m$-th ray at the $k$-th interaction. \\
        $\beta_m^{(k)}$ & Real scalar denoting the outgoing angle of the $m$-th ray at the $k$-th interaction. \\
        $\bm{\zeta}_m^{(k)}$ & Real one-hot vector denoting the scatterer semantic class at the $k$-th interaction of the $m$-th ray. \\
        $\psi_m^{(k)}$ & Real scalar denoting the polarization angle of the $m$-th ray at the $k$-th interaction. \\
        $\gamma_m^{(k)}$ & Real scalar denoting the angular offset between the global and local coordinate systems at the $k$-th interaction. \\
        $\Gamma_{\parallel,m}^{(k)}$, $\Gamma_{\perp,m}^{(k)}$ & Complex scalars denoting the parallel and perpendicular polarization components of the interaction coefficient, respectively. \\
        \midrule
        $\mathcal{E}$ & Structured set denoting the propagation environment, including polygonal scatterer surfaces and semantic classes. \\
        $\mathcal{D}_{tx,i}$, $\mathcal{D}_{rx,j}$ & Tuples denoting the configurations of the $i$-th Tx and the $j$-th Rx, respectively. \\
        $\mathcal{A}_{tx,i}$, $\mathcal{A}_{rx,j}$ & Structured variables denoting Tx/Rx antenna characteristics, including radiation pattern and polarization. \\
        $\mathbf{p}_{tx,i}$, $\mathbf{p}_{rx,j}$ & Real 3D vectors denoting the positions of the $i$-th Tx and the $j$-th Rx, respectively. \\
        \midrule
        $f(\theta)$ & Neural function denoting the outgoing ray prediction network. \\
        $\theta$ & Real vector denoting the trainable parameters of $f(\theta)$. \\
        $\mathcal{S}$ & Set denoting a training or evaluation dataset. \\
        $\mathcal{S}_{\parallel}$, $\mathcal{S}_{\perp}$ & Sets denoting polarization-specific pre-training datasets for parallel and perpendicular components, respectively. \\
        $\mathbf{y}_{\lambda,n}$ & Vector denoting the label of the $n$-th pre-training sample for polarization component $\lambda\in\{\parallel,\perp\}$. \\
        $\phi$ & Mapping denoting the bijection between predicted and label MPC sets. \\
        $\mathcal{L}_{\mathrm{geo}}$ & Real scalar denoting the temporal and angular discrepancy used for MPC association. \\
        \bottomrule
    \end{tabularx}
    \label{Notation}
\end{table*}

In this paper, we propose \textit{Ge}neralizable \textit{Ne}ural \textit{RT} (GeNeRT), a physics-informed approach to ICM that has strong generalization and superior accuracy. While inspired by WiNeRT's core architecture of using a dedicated outgoing ray prediction network for ray-surface interactions \cite{WiNeRT},  GeNeRT distinguishes itself through innovations in the network architecture design and training strategy. 
The main features and novelties of our GeNeRT are summarized as follows:

First, GeNeRT achieves strong generalization by incorporating relative geometric features and scatterer semantics into the ray prediction network. This design reduces spatial dependency as incurred in \cite{WiNeRT} and hence enables reliable multipath component (MPC) prediction across both untrained regions within a scenario and entirely unseen scenarios. In addition, the incorporation of scatterer semantics, namely the essential material features of scatterers that directly affect \gls{EM} wave propagation behavior, 
enhances the representation of the intrinsic characteristics of the overall \gls{EM} behavior.

Second, GeNeRT significantly improves the accuracy of ray-surface interaction modeling through a polarization-driven dual-branch network architecture for outgoing ray prediction. Such dual-branch architecture explicitly models parallel and perpendicular polarization components according to the Fresnel reflection law. It thus effectively captures the polarization-dependent MPC properties. Additionally, we incorporate positional encoding (PosEnc) to process the angular information in the network input, as it enables the network to effectively capture complex nonlinear angular relationships. Furthermore, residual connections (ResNet) are integrated throughout the network to facilitate stable convergence and efficient training of deep structures.

Third, GeNeRT establishes an efficient three-stage training strategy that combines module-wise pre-training, system-wise end-to-end training, and measurement-based fine-tuning. In the first stage, the module-wise pre-training phase uses polarization-specific datasets to guide the outgoing ray prediction network in capturing the universal polarization behavior. In the second stage, the system-wise end-to-end training phase learns site-specific \gls{EM} propagation characteristics based solely on the Rx-side channel impulse responses (CIRs). This eliminates the dependence on full propagation path information, which was required in both WiNeRT and LWDT but is hardly measurable in practice. In the third stage, a measurement-based fine-tuning step adapts the simulation-trained model to real-world environments by refining the polarization-related interaction modules using sparse measured MPCs.

Extensive experiments validate the effectiveness of GeNeRT in outdoor scenarios. Specifically, GeNeRT consistently outperforms state-of-the-art neural RT baselines in three aspects: 1) MPC prediction accuracy within the training region, 2) intra-scenario spatial transferability, and 3) inter-scenario zero-shot generalization capability. Beyond simulation, a measurement-based fine-tuning experiment conducted in a real-world environment further verifies that sparse measured MPCs can effectively adapt GeNeRT to practical propagation characteristics. Furthermore, GeNeRT achieves lower runtime than \gls{WI}~\cite{remcom_wireless_insite}, while ablation studies confirm the necessity of ResNet and the proposed training strategy.

The remainder of this paper is organized as follows. Section II introduces the overall framework of the proposed GeNeRT. Section III details the network architecture and training strategy of the learnable module within GeNeRT. Section IV presents the simulation results. Section V validates GeNeRT through measurement-based fine-tuning experiments. Finally, Section VI concludes the paper. Key notations are listed in Table~\ref{Notation}.

\vspace{-0.85\baselineskip}
\section{Overall Framework}
This section first reviews the basic multipath channel model, then provides an overview of our GeNeRT framework, followed by a description of the neural ray-surface interaction.

\subsection{Multipath Channel Model}
When wireless signals traverse complex environments, they undergo diverse propagation mechanisms, such as free-space propagation, reflection, diffraction, and scattering, leading to multiple paths. Each propagation path introduces amplitude attenuation and phase shifts, collectively shaping the composite signal. The multipath channel is typically represented by a CIR \cite{omnidirectional_CIR}, expressed as
\begin{equation}
    h(\tau, \mathbf{\Theta}, \mathbf{\Phi}) = \sum_{n=1}^{N_{i,j}} a_n \delta(\tau - \tau_n) \delta(\mathbf{\Theta} - \mathbf{\Theta}_n) \delta(\mathbf{\Phi} - \mathbf{\Phi}_n),  \label{CIR}
\end{equation}
where \(\tau_{n}\), \( \mathbf{\Theta}_{n} \), \( \mathbf{\Phi}_{n} \) and $a_{n}$ denote the delay, the azimuth/elevation angle of departure (AoD), the azimuth/elevation AoA, and the total attenuation coefficient of the $n$-th path, respectively. Each $a_{n}$ is determined by the total propagation distance $d_{n}$, the number of interactions $I_{n}$, and the attenuation coefficient at each interaction $\Gamma_{n}^{(k)}$, for $k=1, 2, ..., I_{n}$. It can be expressed as
\begin{equation}
    a_{n} = \underbrace{\frac{\lambda}{4\pi d_{n}}}_{\text{1st}} \cdot \underbrace{\left( \prod_{k=1}^{I_{n}} \Gamma_{n}^{(k)} \right)}_{\text{2nd}} \cdot \underbrace{e^{-\mathrm{j} \frac{2\pi}{\lambda} d_{n}}}_{\text{3rd}}, \label{complex total coeffcient}
\end{equation}
which incorporates three key components: the first term is the free-space attenuation coefficient magnitude, the second term is the attenuation coefficient due to the interaction of the ray with the environment, and the third term represents the phase change caused by free-space propagation. Therefore, multipath channel modeling can be reformulated as the task of predicting a set of MPCs, each characterized by a set  
\begin{equation}
    \mathcal{H}_{n} = \left\{ \tau_{n}, \mathbf{\Theta}_{n}, \mathbf{\Phi}_{n}, a_{n} \right\},
    \label{MPCs Set}
\end{equation}
where \(a_{n}\) depends on \(d_{n}\), \(I_{n}\), and \(\left\{\Gamma_{n}^{(k)}\right\}_{k=1}^{I_{n}}\).

\begin{figure*}[tb]
    \centering
    \includegraphics[width=0.9\linewidth]{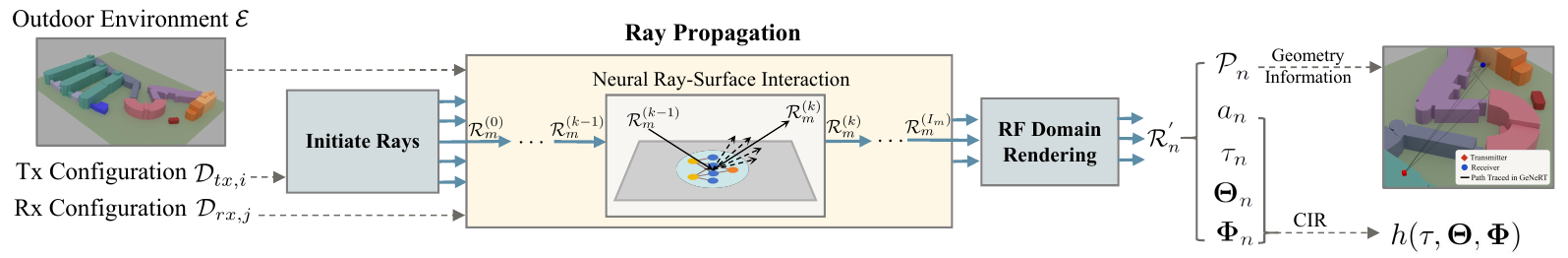}
    \caption{Flowchart of GeNeRT.}
    \label{Flowchart of GEN-RT Framework}
    \vspace{-1.2\baselineskip}
\end{figure*}

\subsection{Overview of GeNeRT Framework}
The core idea of GeNeRT lies in replacing the mathematical equations used to model ray-surface interactions in conventional \gls{RT} with a neural network representation. As illustrated in Fig. \ref{Flowchart of GEN-RT Framework}, rays are emitted from the Tx and propagate through the environment via successive interactions with environmental scatterer surfaces. Each ray is eventually either terminated or received. After collecting all rays that successfully reach the Rx, GeNeRT generates the CIR for the corresponding Tx–Rx pair and provides the geometric information of propagation paths. To be more specific, GeNeRT can be represented by a forward function $\text{GeNeRT}_{\theta}$, which is formulated as
\begin{equation}
    \text{GeNeRT}_{\theta}:(\mathcal{E}, \mathcal{D}_{tx, i}, \mathcal{D}_{rx, j}) \mapsto \left(h(\tau, \mathbf{\Theta, \mathbf{\Phi}}), \left\{ \mathcal{P}_{n}^{(I_{n})}\right\}_{{n}=1}^{N_{i, j}} \right).
    \label{GeNeRT_Function}
\end{equation}
The input to this function consists of propagation environment and Tx/Rx configurations. Concretely, the propagation environment is defined as \( \mathcal{E} \), which consists of a series of convex polygonal surfaces sharing the same number of vertices. Each surface is associated with a one-hot encoded scatterer semantic class, denoted as $\bm{\zeta}$. 
Here, the scatterer semantics refer to the essential material features of scatterers that directly affect \gls{EM} wave propagation behavior, and the scatterer semantic class can be extracted from the physical environment via texture recognition techniques or other appropriate approaches. 
The configurations of the $i$-th Tx and $j$-th Rx, denoted by $\mathcal{D}_{tx, i}$ and $\mathcal{D}_{rx, j}$, respectively, are defined as
\begin{equation}
    \mathcal{D}_{tx, i}=\left\{\mathcal{A}_{tx, i}, \mathbf{p}_{tx,i}\right\}, \:\mathcal{D}_{rx, j}=\left\{\mathcal{A}_{rx, j}, \mathbf{p}_{rx,j}\right\},
\end{equation}
where $\mathcal{A}_{\mathrm{tx}, i}$ ($\mathcal{A}_{\mathrm{rx}, j}$) denotes the antenna characteristics of the Tx (Rx), including the radiation pattern and polarization of the antennas; $\mathbf{p}_{\mathrm{tx},i}$ ($\mathbf{p}_{\mathrm{rx},j}$) represents the spatial coordinates of the Tx (Rx), respectively. 
Based on these inputs, the $\text{GeNeRT}_{\theta}$ in \eqref{GeNeRT_Function} then maps them to $h(\tau, \mathbf{\Theta, \mathbf{\Phi}})$ in (\ref{CIR}) and $\{ \mathcal{P}_{n}^{(I_{n})}\}_{{n}=1}^{N_{i, j}}$, where $N_{i, j}$ denotes the number of propagation paths between the \(i\)-th Tx and the \(j\)-th Rx, $I_{n}$ represents the total number of ray-surface interactions along the ${n}$-th path, $\mathcal{P}_{n}^{(I_{n})} = \{\mathbf{p}_n^{(k)}\}_{k=1}^{I_{n}}$ stores all interaction points along this path, where $\mathbf{p}_n^{(k)}$ denotes the 3D coordinate at the $k$-th interaction of the $n$-th path. 

The implementation process of GeNeRT comprises the following three steps.

\vspace{-0.25\baselineskip}
\subsubsection{Ray Initialization} 

We launch \( M \) different rays from the Tx, each initialized with a state \( \mathcal{R}_m^{(0)}\), as defined later.  
Please be reminded that a ``ray” is any simulated propagation trajectory launched from the Tx, whereas a ``path” is a ray that successfully reaches the Rx to contribute to the channel response. 

\vspace{-0.35\baselineskip}
\subsubsection{Ray Propagation} 
After initialization, rays propagate through the environment and interact with scatterer surfaces. During this process, GeNeRT tracks each ray’s propagation using a ray state. For each ray $m$, the corresponding state at the $k$-th interaction with the environment is defined as
\begin{equation}
    \mathcal{R}_m^{(k)} = \left\{ \mathcal{P}_m^{(k)}, \mathbf{d}_m^{(k)}, \prod_{l=1}^{k} \Gamma_m^{(l)} \right\}, \label{path state}
\end{equation}
where the maximum value of \( k \) is defined as $I_m$ and has two distinct meanings. For rays that successfully reach the Rx, it is defined as the total number of interactions experienced before arrival. For terminated rays, it represents the number of interactions completed prior to termination. The tuple \( \mathcal{P}_m^{(k)} \) stores the spatial coordinates of all interaction points up to the \( k \)-th interaction, and is initialized as an empty tuple when \( k = 0 \). The unit direction vector \( \mathbf{d}_m^{(k)} \) represents the propagation direction of the \( m \)-th ray after the \( k \)-th interaction, with \( \mathbf{d}_m^{(0)} \) denoting the initial direction. The third term corresponds to the cumulative product of the ray's interaction coefficients up to the \( k \)-th interaction, initialized to 1 when \( k = 0 \).

The state in \eqref{path state} is iteratively updated as the ray propagates and interacts with the environment. Specifically, a neural network-based module is employed to predict interaction coefficients and outgoing directions, while geometric calculations and ray state updates are handled through deterministic procedures. Further details are provided in Section II-C.

\vspace{-0.35\baselineskip}
\subsubsection{RF Domain Rendering} 
In general, rendering is the process of transforming abstract data into a form that enables analysis, comprehension, and visualization. In the RF domain, rendering is the process of computing the desired characteristics of RF signals \cite{Pi_Yibo}. To be more specific, taking the $n$-th propagation path as an example, this step serves as the post-processing of the information in $\mathcal{R}_{n}^{(I_{n})}$. On one hand, we preserve the original interaction point coordinates in \( \mathcal{P}_{n}^{(I_{n})} \); on the other hand, we extract the characteristics $\mathcal{H}_{n}$ in (\ref{MPCs Set}), where the parameters $\tau_{n}, \mathbf{\Theta}_{n}, \mathbf{\Phi}_{n}$ are determined by the interaction points $\mathcal{P}_{n}^{(I_{n})}$ along with the positions of the Tx and Rx, while $a_{n}$ can be readily obtained from (\ref{complex total coeffcient}). In summary, the output of this step is denoted as $\mathcal{R}_{n}^{'} = \{\mathcal{P}_{n}^{(I_{n})}, \mathcal{H}_{n} \}$.

\begin{figure*}[t]
    \centering
    \vspace{-0.45\baselineskip}
    \includegraphics[width=0.87\linewidth]{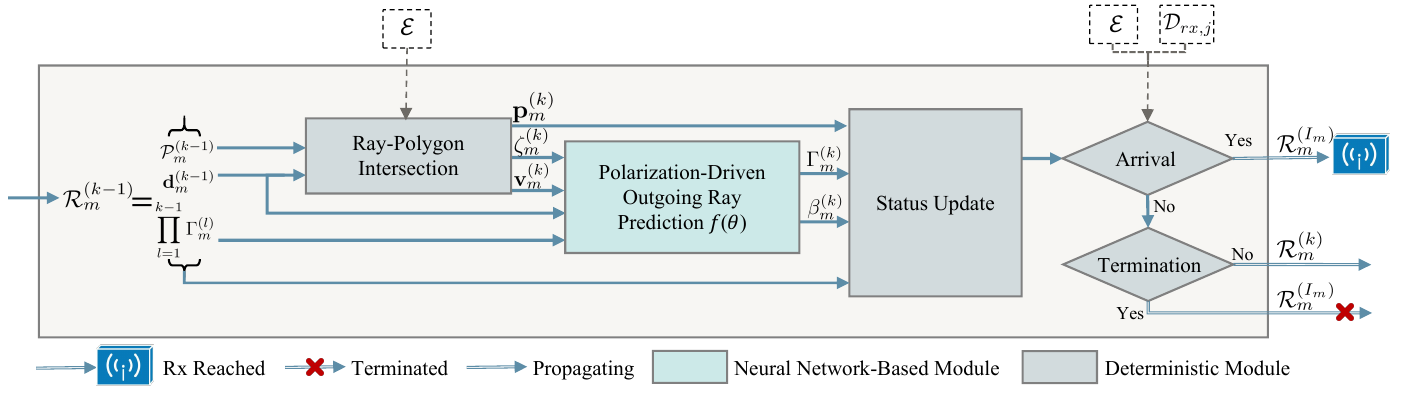}
    \caption{Flowchart of Neural Ray-Surface Interaction.}
    \label{Flowchart of Neural Ray Surface Interaction}
    \vspace{-0.9\baselineskip}
\end{figure*}

\subsection{Neural Ray-Surface Interaction}
\begingroup
\setlength{\abovedisplayskip}{0.12\baselineskip}
\setlength{\belowdisplayskip}{0.12\baselineskip}
\setlength{\abovedisplayshortskip}{0.06\baselineskip}
\setlength{\belowdisplayshortskip}{0.08\baselineskip}

This subsection discusses the detailed implementation process of the ray-surface interaction step, with its flowchart illustrated in Fig. \ref{Flowchart of Neural Ray Surface Interaction}. The ray-surface interaction may in principle involve multiple propagation mechanisms, such as reflection, refraction, scattering, and diffraction. To maintain a clear focus, this work is concentrated on reflection, which we consider foundational to the modeling of other propagation mechanisms. The overall procedure consists of the following four steps.

\vspace{-0.65\baselineskip}
\subsubsection{Ray-Polygon Intersection}
We model the scatterer surfaces via polygonal primitives (e.g., triangle) with an identical number of vertices. Specifically, considering the $k$-th interaction of the $m$-th ray as an example, the corresponding ray equation is formulated as
\begin{equation}
    \mathbf{p}(x) = \mathbf{p}_m^{(k-1)}+x \cdot \mathbf{d}_m^{(k-1)},
\end{equation}
where \(\mathbf{p}(x) \in \mathbb{R}^{3\times1}\) is the coordinate of any point on the ray, and \(x\) is the distance from the point to the starting point. Initially, we solve this equation in conjunction with the plane equations of all surfaces in the environment to compute potential intersection points between the ray and these planes. Subsequently, the Möller–Trumbore algorithm \cite{moller2005fast} is applied to verify whether these intersection points are contained within their corresponding surfaces. Points satisfying this condition are classified as candidate intersection points. Ultimately, the closest candidate point to the ray origin is selected as $\mathbf{p}_m^{(k)}$, representing the spatial coordinates of the $k$-th interaction along the $m$-th ray. Once $\mathbf{p}_m^{(k)}$ is determined, the corresponding polygonal surface is retrieved, from which the scatterer semantic class $\bm{\zeta}_m^{(k)}$ (expressed as a one-hot vector) and the surface normal vector $\mathbf{v}_m^{(k)}$ are obtained. Notably, this step is computation-intensive. To improve its runtime efficiency, we adopt a GPU-accelerated vectorized Möller–Trumbore algorithm, an efficient nearest-intersection filtering strategy, and a fully tensorized computation pipeline executed on GPU.

\vspace{-0.65\baselineskip}
\subsubsection{Polarization-Driven Outgoing Ray Prediction}
In the previous step, we have identified the interaction positions along with their corresponding properties, including the scatterer semantic class and surface normal vector. The current step aims to predict the outgoing direction \(\mathbf{d}_m^{(k)}\) and the interaction coefficient \(\Gamma_m^{(k)}\). While conventional RT frameworks (e.g., Sionna RT \cite{Sionna}) explicitly model this process using the law of reflection and Fresnel equations, our framework employs a neural network to implicitly learn the underlying physical principles to achieve a more accurate and efficient MPC prediction.

In the reflection field, the reflected angle \(\beta_m^{(k)}\) and interaction coefficient \(\Gamma_m^{(k)}\) depend on the incident direction, surface normal, polarization angle, and scatterer semantic class. To reduce spatial dependency and enhance the generalization capability, the network input is designed using relative geometric features instead of absolute coordinates and directions. Specifically, the incident angle \(\alpha_m^{(k)}\) is computed from \(\mathbf{d}_m^{(k-1)}\) and \(\mathbf{v}_m^{(k)}\), while the angular offset \(\gamma_m^{(k)}\) is derived from the polarization angle \(\psi_m^{(k)}\), which quantifies the deviation between the electric field oscillation direction \(\hat{\mathbf{e}}_E\) and the plane of incidence:
\begin{equation}
    \psi_m^{(k)} = \text{arccos}\left( \frac{\mathbf{v}_m^{(k)} \times \mathbf{d}_m^{(k-1)}}{||\mathbf{v}_m^{(k)} \times \mathbf{d}_m^{(k-1)}||} \cdot \hat{\mathbf{e}}_E \right). \label{Polatization angle}
\end{equation}
This angle plays a crucial role in determining interaction loss, primarily influenced by two factors. The first is the angular deviation between the global and local coordinate systems, corresponding to \(\gamma_m^{(k)}\), which can be computed by assuming \(\hat{\mathbf{e}}_E\) is vertically upward in \eqref{Polatization angle}.
The second is the orientation of \(\hat{\mathbf{e}}_E\), which depends on the ray’s historical interactions and is difficult to model explicitly. To approximate this effect, we use the cumulative product of the previous interaction coefficients, denoted as \(\prod_{l=1}^{k-1} \Gamma_m^{(l)}\). This design allows the network to accurately capture the polarization-related behaviors. Accordingly, the polarization-driven outgoing ray prediction network is defined as
\begin{equation}
    f: (\alpha_m^{(k)}, \bm{\zeta}_m^{(k)}, \gamma_m^{(k)}, \prod_{l=1}^{k-1} \Gamma_m^{(l)};\theta) \mapsto (\beta_m^{(k)}, \Gamma_m^{(k)}), \label{Network function}
\end{equation}
where \(\theta\) denotes the trainable parameters.

\vspace{-1.25\baselineskip}
\subsubsection{Status Update}
At each ray-surface interaction, the state is updated to reflect new propagation conditions. Specifically, for the \(k\)-th interaction along the \(m\)-th ray, the predicted reflection field yields the reflected angle \(\beta_m^{(k)}\) and interaction coefficient \(\Gamma_m^{(k)}\), which are used to update the ray state. The updated state $\mathcal{R}_m^{(k)}$, derived from the previous state $\mathcal{R}_m^{(k-1)}$, can be expressed as
\begin{equation}
\begin{aligned}
    \mathcal{R}_m^{(k)}
    &= \text{Update}(\mathcal{R}_m^{(k-1)}) \\
    &= \big\{ \mathcal{P}_m^{(k-1)}\oplus\mathbf{p}_m^{(k)}, \mathbf{d}_m^{(k)}, \Gamma_m^{(k)} \cdot \prod_{l=1}^{k-1} \Gamma_m^{(l)} \big\},
\end{aligned}
\label{updating path state}
\end{equation}
where $\oplus$ denotes the tuple concatenation operation, and the outgoing direction $\mathbf{d}_m^{(k)}$ can be calculated as
\begin{equation}
\begin{aligned}
    \mathbf{d}_m^{(k)}
    &= \cos \beta_m^{(k)} \cdot \mathbf{v}_m^{(k)} \\
    &\quad + \sin \beta_m^{(k)} \cdot
    \frac{\mathbf{d}_m^{(k-1)} - (\mathbf{d}_m^{(k-1)} \cdot \mathbf{v}_m^{(k)}) \mathbf{v}_m^{(k)}}
    {\| \mathbf{d}_m^{(k-1)} - (\mathbf{d}_m^{(k-1)} \cdot \mathbf{v}_m^{(k)}) \mathbf{v}_m^{(k)} \|}.
\end{aligned}
\end{equation}

\vspace{-1.25\baselineskip}
\subsubsection{Arrival/Termination Judgment}

Following the computation of the outgoing rays and their associated states, each ray undergoes a sequential evaluation to determine whether it (i) reaches a Rx, (ii) continues to propagate with an updated state, or (iii) terminates because of environmental boundaries, a predefined maximum number of interactions, or a predefined power threshold.

\endgroup
\section{Polarization-Driven Outgoing Ray Prediction}
The core component of GeNeRT is the outgoing ray prediction network \( f(\theta) \), which plays a critical role in accurately predicting MPCs and interaction characteristics. This section begins with an introduction to the polarization-aware reflection model, then describes the network architecture of \( f(\theta) \), and concludes with the training strategy employed for the network.

\subsection{Polarization-Aware Reflection Model}
\begin{figure}[t]
    \centering
    \includegraphics[width=0.85\linewidth]{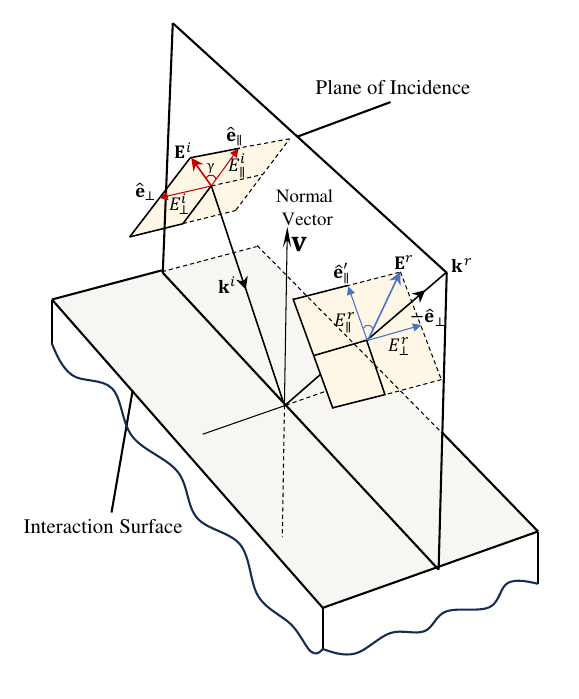}
    \caption{Schematic diagram of the decomposition of the incident and reflected waves in parallel and perpendicular polarization direction. \( \mathbf{k}^i \) and \( \mathbf{k}^r \) represent the propagation directions of the incident and reflected waves, respectively.}
    \label{E-field decomposition}
\end{figure}

The outgoing ray prediction network is designed around reflection physics rather than as an unconstrained attenuation regressor. Since the reflected field depends on how the incident electric field decomposes relative to the plane of incidence, we first introduce the Fresnel reflection relation that motivates the subsequent dual-branch polarization design. The Fresnel reflection law rigorously defines the reflection attenuation coefficient, which characterizes the signal strength and field distribution of the reflected wave and is directly related to the accuracy of reflected field prediction. For simplicity, the reflection coefficient \( \Gamma_n^{(k)} \) is denoted as \( \Gamma^r \) in the following derivations. It is defined as
\begin{equation}
    \Gamma^r = \frac{\mathbf{E}^r}{\mathbf{E}^i}, \label{reflection attenuation coefficient definition}
\end{equation}
where \(\mathbf{E}^r\) and \(\mathbf{E}^i\) represent the reflected and incident electric field intensities, respectively. However, directly modeling the electric field in 3D space is inherently complex, requiring its decomposition into components perpendicular and parallel to the plane of incidence \cite{rappaport2024wireless}. Fig. \ref{E-field decomposition} illustrates the decomposition details of the \gls{EM} wave before and after reflection. The incident electric field \(\mathbf{E}^i\) can be decomposed as
\begin{equation}
    \mathbf{E}^i = E^i_{\perp} \cdot \hat{\mathbf{e}}_\perp + E^i_{\parallel} \cdot \hat{\mathbf{e}}_\parallel,
\end{equation}
where \(\hat{\mathbf{e}}_\perp\) and \(\hat{\mathbf{e}}_\parallel\) denote the unit vectors in the directions perpendicular and parallel to the plane of incidence, respectively. Given the polarization angle \(\psi\), the perpendicular and parallel components of the incident electric field, \(E_{\perp}^{i}\) and \(E_{\parallel}^{i}\), can be computed as

\begin{equation}
    E_{\perp}^{i} = |\mathbf{E}^{i}| \cdot \sin(\psi), \quad E_{\parallel}^{i} = |\mathbf{E}^{i}| \cdot \cos(\psi).
\end{equation}
Similarly, the reflected electric field can also be expressed as
\begin{equation}
    \mathbf{E}^r = E^r_{\perp} \cdot \hat{\mathbf{e}}_\perp + E^r_{\parallel} \cdot \hat{\mathbf{e}}^{'}_\parallel, \label{decomposition of reflected field}
\end{equation}
where \(\hat{\mathbf{e}}^{'}_\parallel\) represents the unit vector in the direction parallel to the plane of incidence for the reflected wave. According to \cite{rappaport2024wireless}, the relationship between the reflected and incident field components is given by
\begin{equation}
\begin{bmatrix}
E_{\perp}^{r} \\
E_{\parallel}^{r}
\end{bmatrix}
=
\begin{bmatrix}
\Gamma_{\perp} & 0 \\
0 & \Gamma_{\parallel}
\end{bmatrix}
\cdot
\begin{bmatrix}
\text{sin}(\psi) \\
\text{cos}(\psi)
\end{bmatrix}\cdot|\mathbf{E}^{i}|,  \label{reflection field computation}
\end{equation}
where \(\Gamma_\perp\) and \(\Gamma_\parallel\) are the perpendicular and parallel polarization components of the reflection coefficient $\Gamma^r$, respectively. These components depend solely on the angle of incidence and the material properties \cite{rappaport2024wireless}.

In summary, the reflection coefficient \(\Gamma^r\) is obtained in two stages: first, computing each polarization component based on material properties and incident angle; second, weighting them by the sine and cosine of the polarization angle \(\psi\).

\subsection{Network Architecture}

\begin{figure*}[t]
    \centering
    \includegraphics[width=0.8\linewidth]{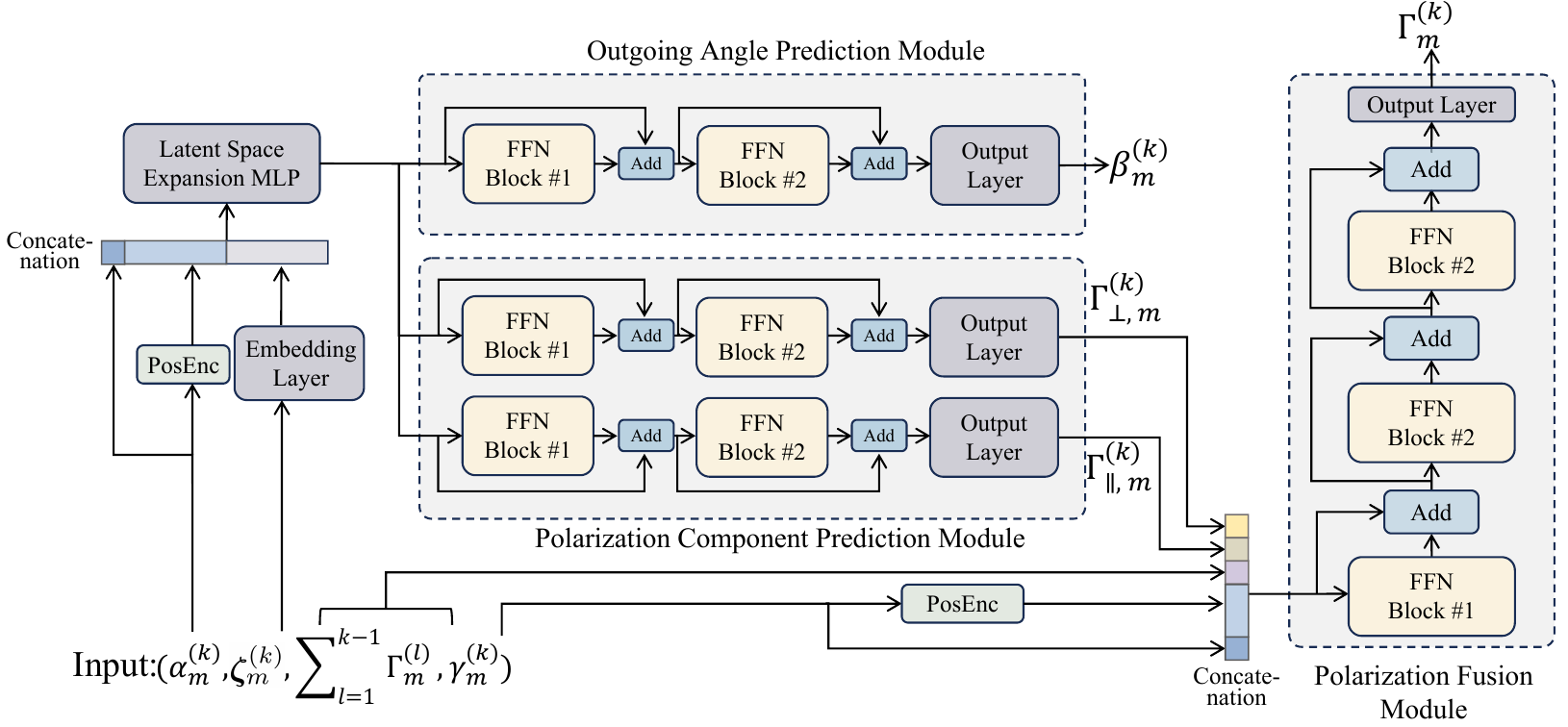}
    \caption{Network architecture for the proposed polarization-driven outgoing ray prediction network $f(\theta)$.}
    \label{NN_Structure}
\end{figure*}

The architecture of the proposed outgoing ray prediction network \(f(\theta)\) is illustrated in Fig.~\ref{NN_Structure}. It consists of three functionally coordinated modules: the outgoing angle prediction module, the polarization component prediction module, and the polarization fusion module, each designed to capture distinct but complementary aspects of EM ray-surface interactions.

The reflection response can vary nonlinearly with angular variables, especially near grazing incidence or when different polarization components dominate. Directly feeding a scalar angle to an MLP may limit its ability to represent such high-frequency angular variations. Inspired by sinusoidal positional encoding and Fourier feature mappings~\cite{mildenhall2020nerf,tancik2020fourier}, we encode each angular input \(x\) by the PosEnc defined in
\begin{equation}
    \begin{aligned}
    \text{PosEnc}(x) = [&\sin(2^1 x), \cos(2^1 x), \ldots, \\
    &\sin(2^{L}x), \cos(2^{L}x)]^T,\quad x\in (0,\pi/2).
    \end{aligned}
    \label{PosEnc}
\end{equation}
In this work, \(x\) denotes either the incident angle \(\alpha_m^{(k)}\) or the angular offset \(\gamma_m^{(k)}\), and \(L\) controls the highest encoded angular frequency.

The network first takes as input the incident angle \(\alpha_m^{(k)}\) and the scatterer semantic class \(\bm{\zeta}_m^{(k)}\). The former is encoded by~\eqref{PosEnc}. The latter is embedded into a continuous latent vector via an embedding layer. These two encoded inputs are concatenated and transformed by a latent space expansion MLP to form a unified feature vector. This feature is then processed by two parallel branches. The outgoing angle prediction module estimates \(\beta_m^{(k)}\) using a stack of feed-forward network (FFN) blocks with ResNet. In parallel, the polarization component prediction module employs a Fresnel-inspired dual-branch architecture to separately compute the parallel \(\Gamma_{\parallel,m}^{(k)}\) and perpendicular \(\Gamma_{\perp,m}^{(k)}\) components of $\Gamma_{m}^{(k)}$, each through its own FFN-based pathway and dedicated output layer. Finally, the predicted polarization components, along with the cumulative coefficient \(\sum_{l=1}^{k-1} \Gamma_m^{(l)}\) and the angular offset \(\gamma_m^{(k)}\) (encoded via PosEnc in~\eqref{PosEnc}), are concatenated and passed to the polarization fusion module, which refines the combined input through stacked FFN blocks and outputs the final interaction coefficient \(\Gamma_m^{(k)}\).

\vspace{-0.75\baselineskip}
\subsection{Training Strategy}
The training procedure of the network \( f(\theta) \) consists of three stages: module-wise pre-training, system-wise end-to-end training, and measurement-based fine-tuning. The first two stages are conducted in simulation to learn general ray-surface interaction behavior and site-specific propagation characteristics, while the third stage uses sparse measured MPCs to adapt the simulation-trained model to real-world environments. Before elaborating on these training stages, we first introduce the dataset with polarization components.

\subsubsection{Polarization-Specific Dataset Construction}
When \gls{EM} waves interact with the environment, they typically contain both parallel and perpendicular polarization components simultaneously. Directly using such data for training introduces two challenges: first, it demands highly precise polarization angles; second, the underlying patterns the network needs to capture become overly complex. Therefore, it is crucial to construct datasets containing only a single polarization component (either parallel or perpendicular) for initial training, enabling the network to effectively learn characteristics of each polarization component. We denote the dataset with parallel and perpendicular polarization components as $\mathcal{S}_{\parallel}$ and $\mathcal{S}_{\perp}$. To precisely control the target link to contain exclusively perpendicular or parallel polarization components, we select only single-reflection paths and employ specialized configurations of Tx and Rx. To be more precise, by employing a particular Tx polarization mode and carefully arranging the positions of Tx and Rx, we ensure that the polarization angle \(\psi\) of the incident \gls{EM} waves on single-reflection paths is either $0$ or $\pi/2$. Specifically, paths with a polarization angle of $0$ are leveraged to construct dataset $\mathcal{S}_{\parallel}$, while those with a polarization angle of $\pi/2$ are utilized to form dataset $\mathcal{S}_{\perp}$. The explicit formulations of $\mathcal{S}_{\parallel}$ and $\mathcal{S}_{\perp}$ are given as
\begin{equation}
    \mathcal{S}_\lambda = \left\{ (\mathbf{x}_n, \mathbf{y}_{\lambda, n}) \right\}_{n=1}^{N_{\text{samples}}},
\end{equation}
where $N_{\text{samples}}$ denotes the number of samples, and $\lambda \in \{\perp, \parallel\}$. The input $\mathbf{x}_n$ and corresponding label $\mathbf{y}_{\lambda,n}$ are explicitly defined as
\begin{equation}
    \mathbf{x}_n = [\bm{\zeta}_n^{(1)}, \alpha_n^{(1)}], \quad \mathbf{y}_{\lambda, n} = [\beta_n^{(1)}, \Gamma_{\lambda, n}^{(1)}],  \label{pre-training feature}
\end{equation}
where $\bm{\zeta}_n^{(1)}$, $\alpha_n^{(1)}$, $\beta_n^{(1)}$ and $\Gamma_{\lambda, n}^{(1)}$ are the scatterer semantic class (one-hot encoding), incident angle, outgoing angle and polarization-specific component of attenuation coefficient for each single-reflection path.

\subsubsection{Module-Wise Pre-Training}
During the module-wise pre-training phase, we leverage $\mathcal{S}_{\parallel}$ and $\mathcal{S}_{\perp}$ to initialize the embedding layer, latent space expansion MLP, polarization component prediction module, and outgoing angle prediction module. This phase is performed offline using simulation-generated single-reflection ray samples, where the incident angle, outgoing angle, and polarization-specific interaction coefficient are available from the ray geometry. Specifically, we adopt a greedy layer-wise training strategy. First, $\Gamma_{\parallel,n}^{(1)}$ in $\mathcal{S}_{\parallel}$ is used to train the embedding layer, the latent space expansion MLP, and the lower branch of the polarization component prediction module. Next, we freeze their weights and use $\Gamma_{\perp,n}^{(1)}$ in $\mathcal{S}_{\perp}$ to train the upper branch of the polarization component prediction module. Finally, we freeze the parameters of all previously trained modules and use the label $\beta_n^{(1)}$ from $\mathcal{S}_{\parallel}$ and $\mathcal{S}_{\perp}$ to train the outgoing angle prediction module. When predicting the polarization components of $\Gamma_m^{(k)}$, we employ \gls{NMSE} as the loss function; when predicting the outgoing angle, we represent each angle using its sine--cosine embedding, $(\sin\beta, \cos\beta)$. The embedding-space \gls{MSE} between the predicted and ground-truth angles is then defined as
\begin{equation}
    \text{MSE} = (\sin\hat{\beta}-\sin{\beta})^2 + (\cos\hat{\beta}-\cos{\beta})^2,
\end{equation}
where $\hat{\beta}$ is the predicted outgoing angle. It is worth noting that the above ray-level labels are used only for offline module-wise pre-training. In the subsequent system-wise end-to-end training phase, GeNeRT does not require ground-truth AoD, interaction points, or full propagation-path labels; delay and AoA are used only for MPC association, while the received attenuation coefficient provides the training supervision.

\subsubsection{System-Wise End-to-End Training}

During the system-wise end-to-end training phase, we use the CIR of each Tx–Rx pair to further train \( f(\theta) \). At this point, the configurations of Tx and Rx are no longer meticulously arranged; instead, they are adjusted based on practical requirements, and the propagation paths may involve multiple interactions. The dataset at this stage can be written as
\begin{equation}
    \mathcal{S} = \{ (\mathcal{X}_{i,j}, \mathcal{Y}_{i,j}) \mid i = 1, \dots, N_t, \; j = 1, \dots, N_r \},
\end{equation}
where \(N_t\) and \(N_r\) are the numbers of the Tx and Rx, respectively. The input \(\mathcal{X}_{i,j}\) and corresponding label \(\mathcal{Y}_{i,j}\) are defined as
\begin{equation}
    \mathcal{X}_{i,j} = \{\mathcal{E}, \mathcal{D}_{tx, i}, \mathcal{D}_{rx, j}\}, 
    \quad 
    \mathcal{Y}_{i,j} = \{ \mathcal{H}_n \}_{n=1}^{N_{i, j}},  \label{end-to-end feature}
\end{equation}
where \(N_{i,j}\) denotes the number of propagation paths between the \(i\)-th Tx and the \(j\)-th Rx. Prior to training the network, we utilize the time delay $\tau_n$ and AoA $ \mathbf{\Phi}_n $ in \(\mathcal{H}_n\) to establish a bijection \(\phi\) between the predicted set $\hat{\mathcal{Y}}_{i,j}$ and the label set ${\mathcal{Y}}_{i,j}$. The predicted set is defined as
\begin{equation}
    \hat{\mathcal{Y}}_{i,j} = \{ \hat{\mathcal{H}}_n\}_{n=1}^{\hat{N}_{i, j}},
\end{equation}
where $\hat{N}_{i, j}$ is the predicted number of paths between the $i$-th Tx and the $j$-th Rx. To be more specific, bijection \(\phi\) is determined as
\begin{equation}
\min_{\phi:\hat{\mathcal{Y}}_{i, j} \to {\mathcal{Y}}_{i, j}} \sum_{\hat{\mathcal{H}}_n\in\hat{\mathcal{Y}}_{i, j}}\mathcal{L}_{\text{geo}}\bigl(\hat{\mathcal{H}_n}, \phi(\hat{\mathcal{H}}_n)\bigr),  \label{Matching pre and label}
\end{equation}
where $\mathcal{L}_{\text{geo}}\bigl(\hat{\mathcal{H}}, \phi(\hat{\mathcal{H}})\bigr)$ measures the temporal and spatial discrepancy between the sets, and is defined as
\begin{equation}
\mathcal{L}_{\text{geo}}(\hat{\mathcal{H}}, \phi(\hat{\mathcal{H}})) = 
\sum_{x \in \{\tau_n,\,\Phi_n\}} 
\frac{ \left\| \hat{x} - x \right\|_2^2 }{ \left\| x \right\|_2^2 }. \label{geometry error}
\end{equation}
Afterward, we use the bijection \(\phi\) to assign label values to the predicted attenuation coefficients and take their \gls{NMSE} as the loss function during training. During this phase, we focus on optimizing specific modules of the network. Unless otherwise specified, only the weights of the polarization fusion module are activated during training.

\subsubsection{Measurement-Based Fine-Tuning}
After module-wise pre-training and system-wise end-to-end training, measurement-based fine-tuning serves as the final training stage to reduce the gap between simulation and real-world propagation measurements. Starting from the simulation-trained network, we unfreeze the weights of the polarization component prediction module and the polarization fusion module, while keeping the remaining modules frozen. This design preserves the geometric propagation prior learned from simulation and focuses the adaptation on the modules that directly characterize ray-surface \gls{EM} interactions.

Given the measured MPC set for each Tx--Rx pair, we first run the simulation-trained GeNeRT model to obtain the predicted MPC set. Since the predicted MPCs and the measured MPCs are unordered and do not naturally have one-to-one correspondence, we employ the matching strategy in \eqref{Matching pre and label} to establish their association based on temporal and angular consistency. The matched measured MPCs are then used as supervision for the corresponding predicted MPCs. Finally, the polarization component prediction module and the polarization fusion module are fine-tuned by minimizing the \gls{NMSE} between the predicted and measured attenuation coefficients. In this way, the model retains the physically informed structure learned from simulation while adapting its \gls{EM} interaction parameters to real measurements.

\section{Simulation Results}

In this section, we evaluate the proposed GeNeRT framework through simulation-based experiments. 
Section IV-A introduces the simulation scenario and dataset configuration, while Section IV-B describes the network architecture, training hyperparameters, and evaluation metrics. The performance within the training area is reported in Section IV-C, followed by the generalization evaluation in Section IV-D. 
Finally, Section IV-E discusses the runtime efficiency of GeNeRT and examines the key network components through ablation studies.

\begin{table*}[!tb]
\centering
\caption{Overview of constructed datasets.}
\begin{threeparttable}
\footnotesize
\setlength{\tabcolsep}{2.4pt}
\renewcommand\arraystretch{1.15}
\begin{tabularx}{0.98\textwidth}{>{\centering\arraybackslash}m{2.45cm}|
                >{\centering\arraybackslash}m{1.25cm}|
                >{\centering\arraybackslash}m{1.25cm}|
                >{\centering\arraybackslash}m{1.55cm}|
                >{\centering\arraybackslash}X|
                >{\centering\arraybackslash}m{1.1cm}|
                >{\centering\arraybackslash}m{1.1cm}|
                >{\centering\arraybackslash}m{1.45cm}}
\hline
\textbf{} & \textbf{Dataset} & \textbf{Scenario} & \makecell{\textbf{Humidity}\\\textbf{Level}} & \makecell{\textbf{Horizontal Positions}\\\textbf{of Rxs}} & \makecell{\textbf{Rx}\\\textbf{Height}} & \makecell{\textbf{Rx}\\\textbf{Count}} & \makecell{\textbf{Train/Val/Test}\\\textbf{Ratio}} \\
\hline
\multirow{2}{*}{\shortstack{Module-Wise\\Pre-Training}}
& $\mathcal{S}_{\perp}$ & Scenario 1 & Dry & Polygon Layout 1 in Fig.~\ref{fig:combined_scene_views}(b) & 2 m & 13500 & 80\:/\:20\:/\:0 \\
& $\mathcal{S}_{\parallel}$ & Scenario 1 & Dry & Polygon Layout 1 in Fig.~\ref{fig:combined_scene_views}(b) & 2 m & 13500 & 80\:/\:20\:/\:0 \\
\hline
\multirow{3}{*}{\shortstack{System-Wise\\End-to-End Training}}
& $\mathcal{S}_{1}$ & Scenario 2 & Dry & Polygon Layout 2 in Fig.~\ref{fig:combined_scene_views}(d) & 6 m & 2029 & 80\:/\:10\:/\:10 \\
& $\mathcal{S}_{2}$ & Scenario 2 & Medium Dry & Polygon Layout 2 in Fig.~\ref{fig:combined_scene_views}(d) & 6 m & 2029 & 80\:/\:10\:/\:10 \\
& $\mathcal{S}_{3}$ & Scenario 2 & Wet & Polygon Layout 2 in Fig.~\ref{fig:combined_scene_views}(d) & 6 m & 2029 & 80\:/\:10\:/\:10 \\
\hline
\multirow{4}{*}{\shortstack{Generalization\\Test}}
& $\mathcal{S}_{4}$ & Scenario 2 & Dry & Polygon Layout 3 in Fig.~\ref{fig:combined_scene_views}(d) & 6 m & 2520 & 0\:/\:0\:/\:100 \\
& $\mathcal{S}_{5}$ & Scenario 2 & Dry & Route Layout 1 in Fig.~\ref{fig:combined_scene_views}(d) & 6 m & 1431 & 0\:/\:0\:/\:100 \\
& $\mathcal{S}_{6}-\mathcal{S}_{14}$ & Scenario 2 & Dry & Polygon Layout 2 in Fig.~\ref{fig:combined_scene_views}(d) & varied\textsuperscript{*} & 2029 & 0\:/\:0\:/\:100 \\
& $\mathcal{S}_{15}$ & Scenario 3 & Dry & Polygon Layout 4 in Fig.~\ref{fig:combined_scene_views}(f) & 6 m & 2352 & 0\:/\:0\:/\:100 \\
\hline
\end{tabularx}
\begin{tablenotes}
\footnotesize
\item[*] Rx heights for $\mathcal{S}_6$ to $\mathcal{S}_{14}$ are set to 2\,m, 4\,m, 8\,m, 10\,m, 12\,m, 14\,m, 16\,m, 18\,m, and 20\,m, respectively.
\end{tablenotes}
\end{threeparttable}
\label{WI Settings}
\vspace{-0.35\baselineskip}
\end{table*}

\subsection{Simulation Data Generation}
In this paper, we leverage \gls{WI} to generate labeled data. \gls{WI} is a commonly used RT software that can generate CIRs. It can trace the rays propagating in the environment and yield the corresponding channel characteristics.

\subsubsection{Scenario Setup}
\begin{figure*}[p]
    \centering

    \begin{subfigure}[t]{0.438\linewidth}
        \centering
        \includegraphics[width=\linewidth]{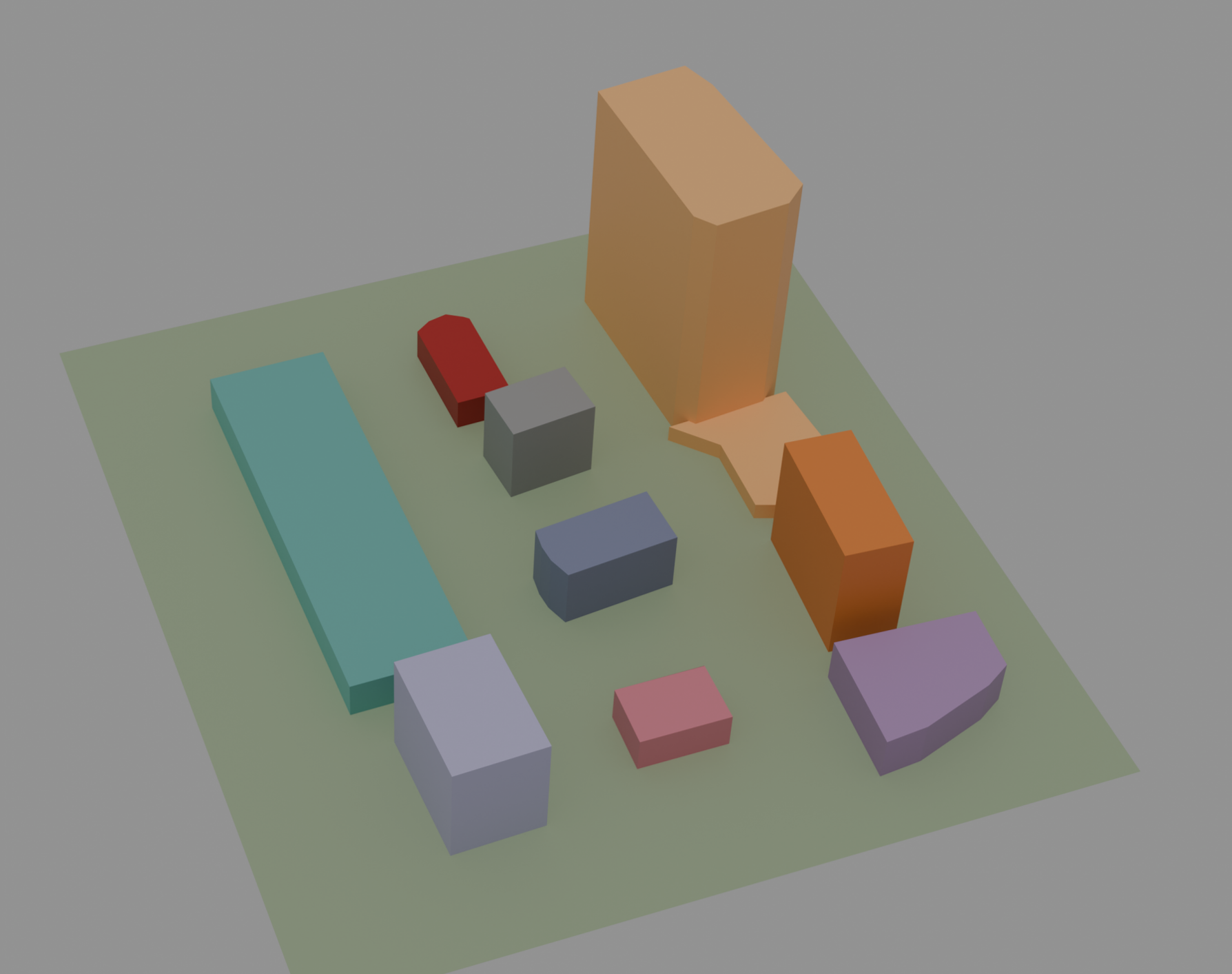}
        \caption{Aerial view of the buildings in Scenario 1.}
    \end{subfigure}
    \hspace{0.04\linewidth}
    \begin{subfigure}[t]{0.438\linewidth}
        \centering
        \includegraphics[width=\linewidth]{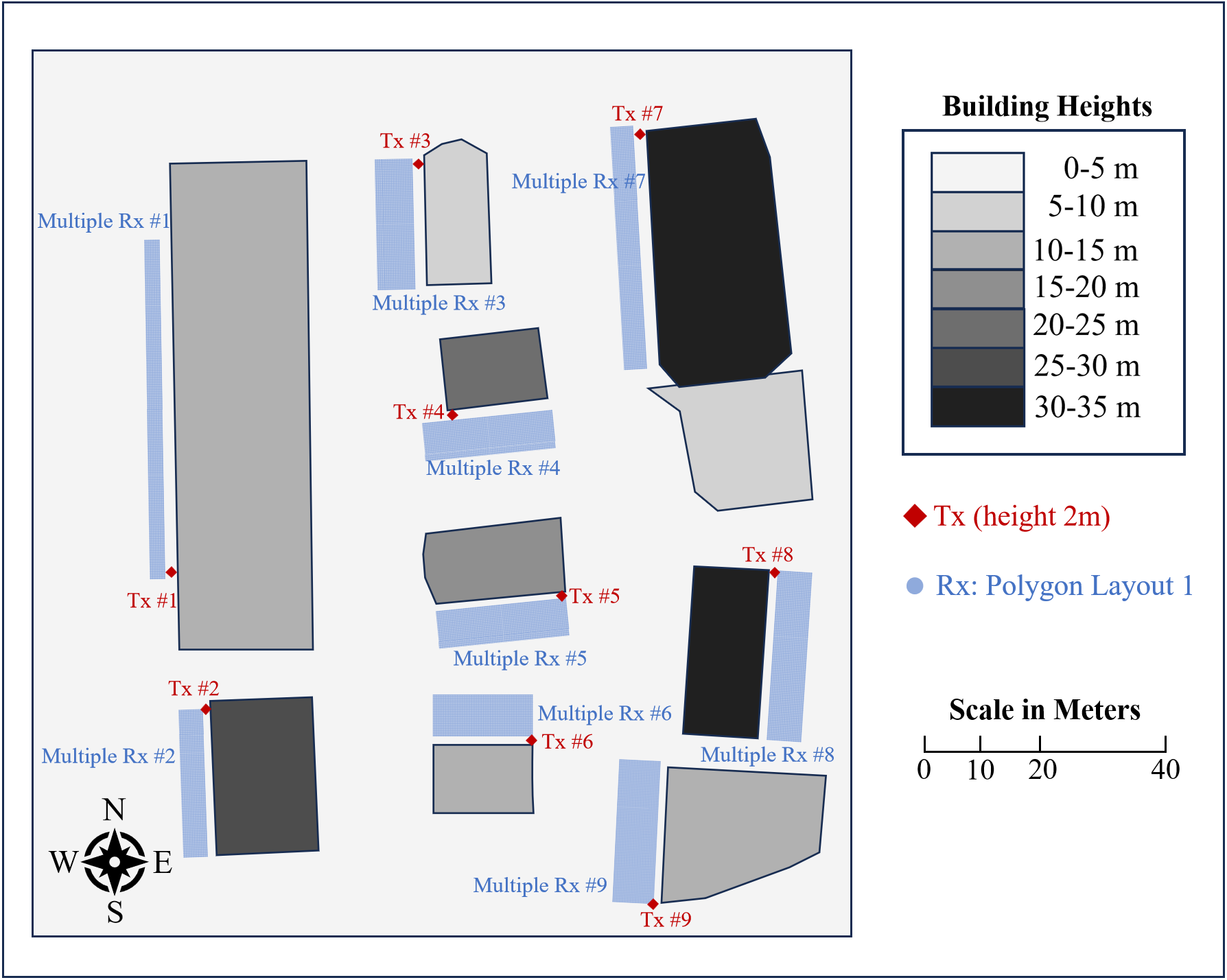}
        \caption{Bird’s eye view of Scenario 1 and the horizontal positions of Txs and Rxs.}
    \end{subfigure}

    \vspace{0pt}

    \begin{subfigure}[t]{0.438\linewidth}
        \centering
        \includegraphics[width=\linewidth]{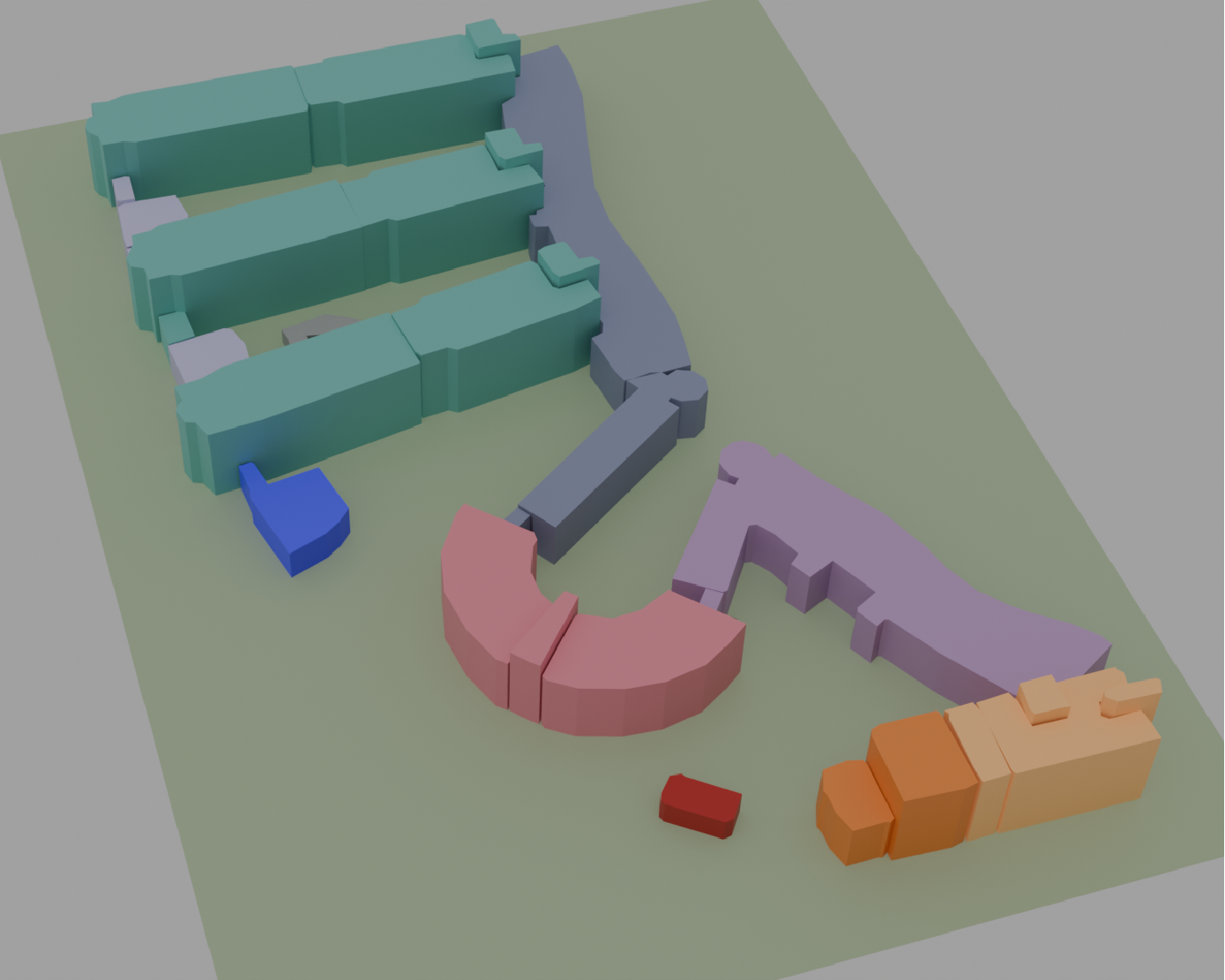}
        \caption{Aerial view of the buildings in Scenario 2.}
    \end{subfigure}
    \hspace{0.04\linewidth}
    \begin{subfigure}[t]{0.438\linewidth}
        \centering
        \includegraphics[width=\linewidth]{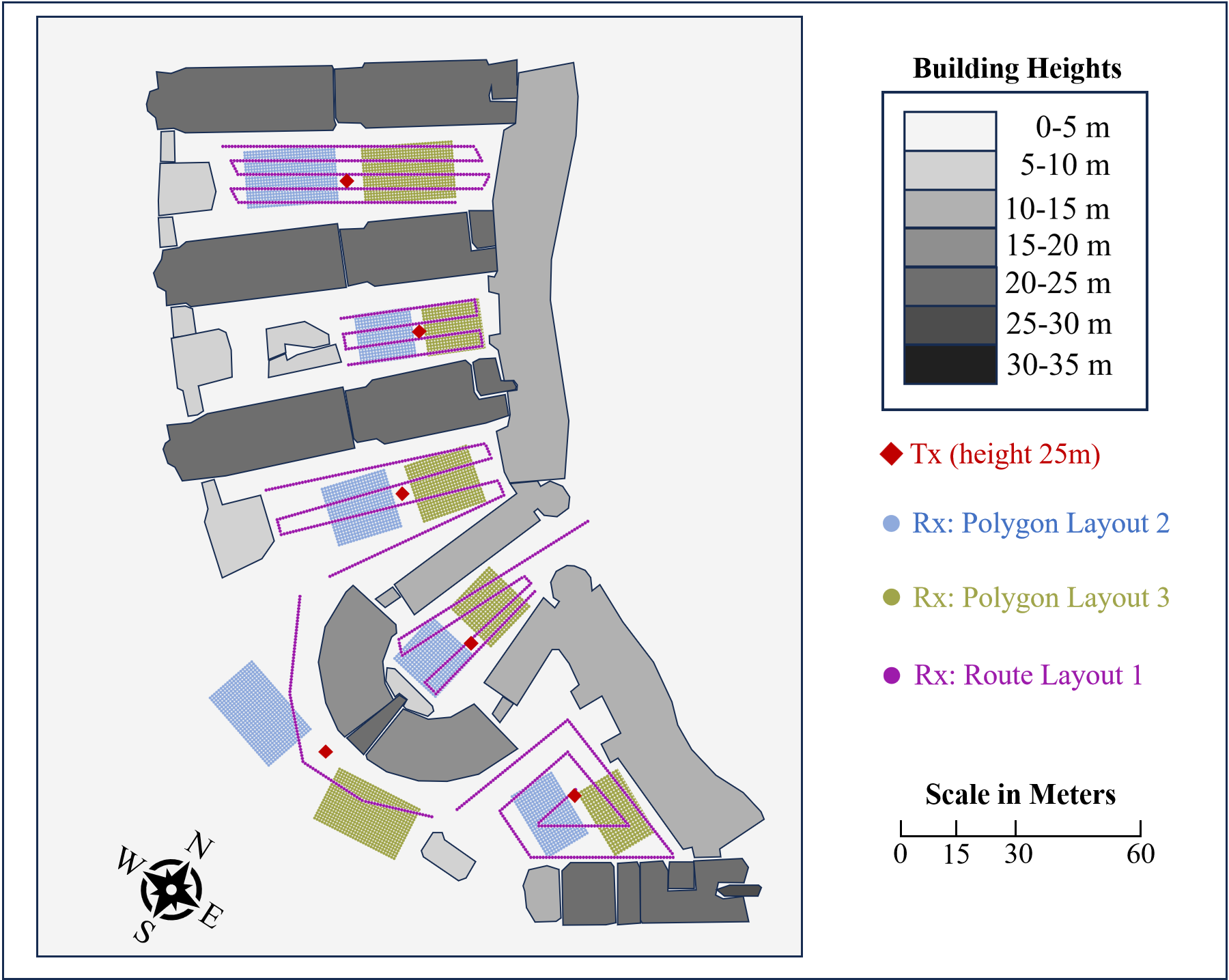}
        \caption{Bird’s eye view of Scenario 2 and the horizontal positions of Txs and Rxs.}
    \end{subfigure}

    \vspace{0pt}

    \begin{subfigure}[t]{0.438\linewidth}
        \centering
        \includegraphics[width=\linewidth]{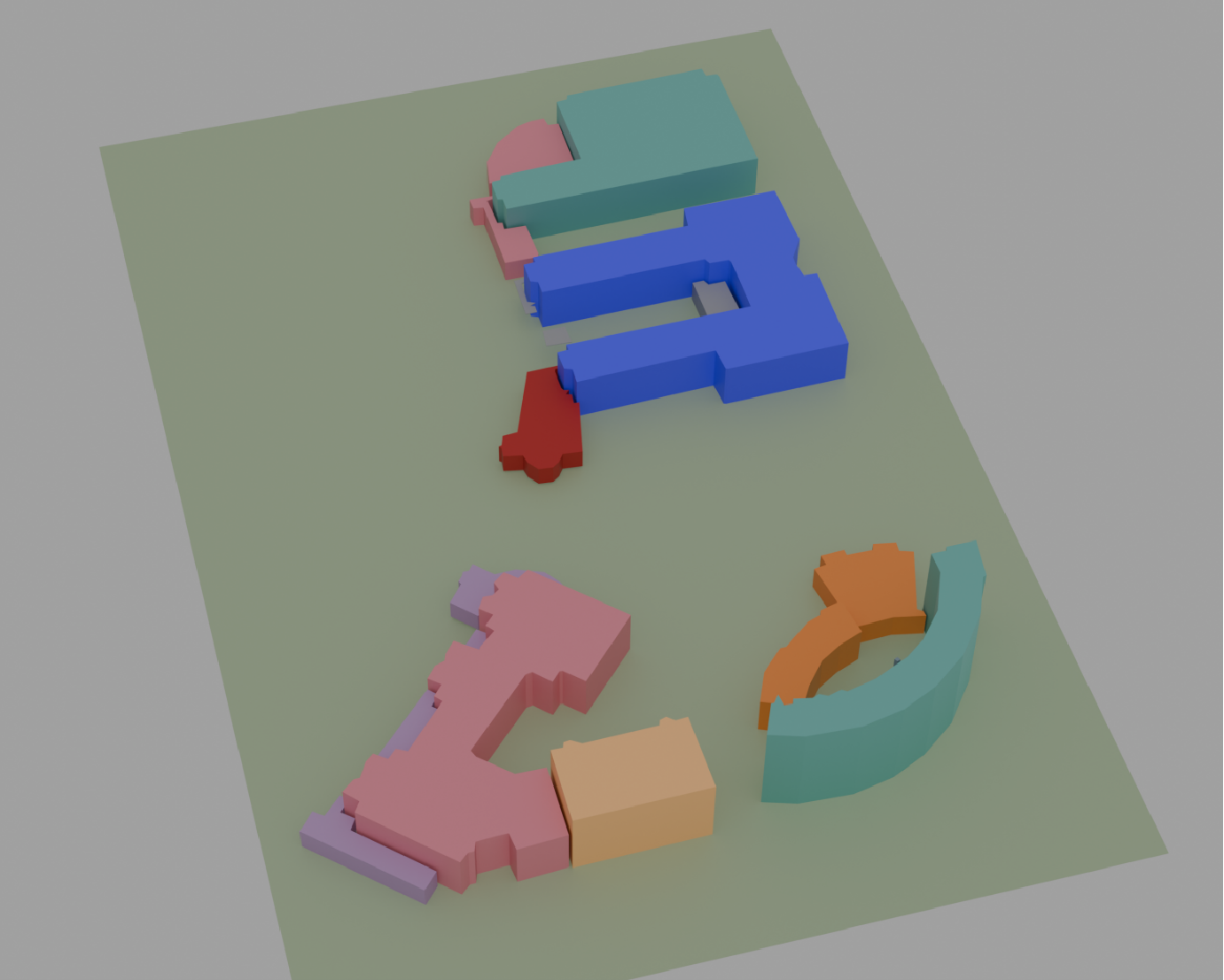}
        \caption{Aerial view of the buildings in Scenario 3.}
    \end{subfigure}
    \hspace{0.04\linewidth}
    \begin{subfigure}[t]{0.438\linewidth}
        \centering
        \includegraphics[width=\linewidth]{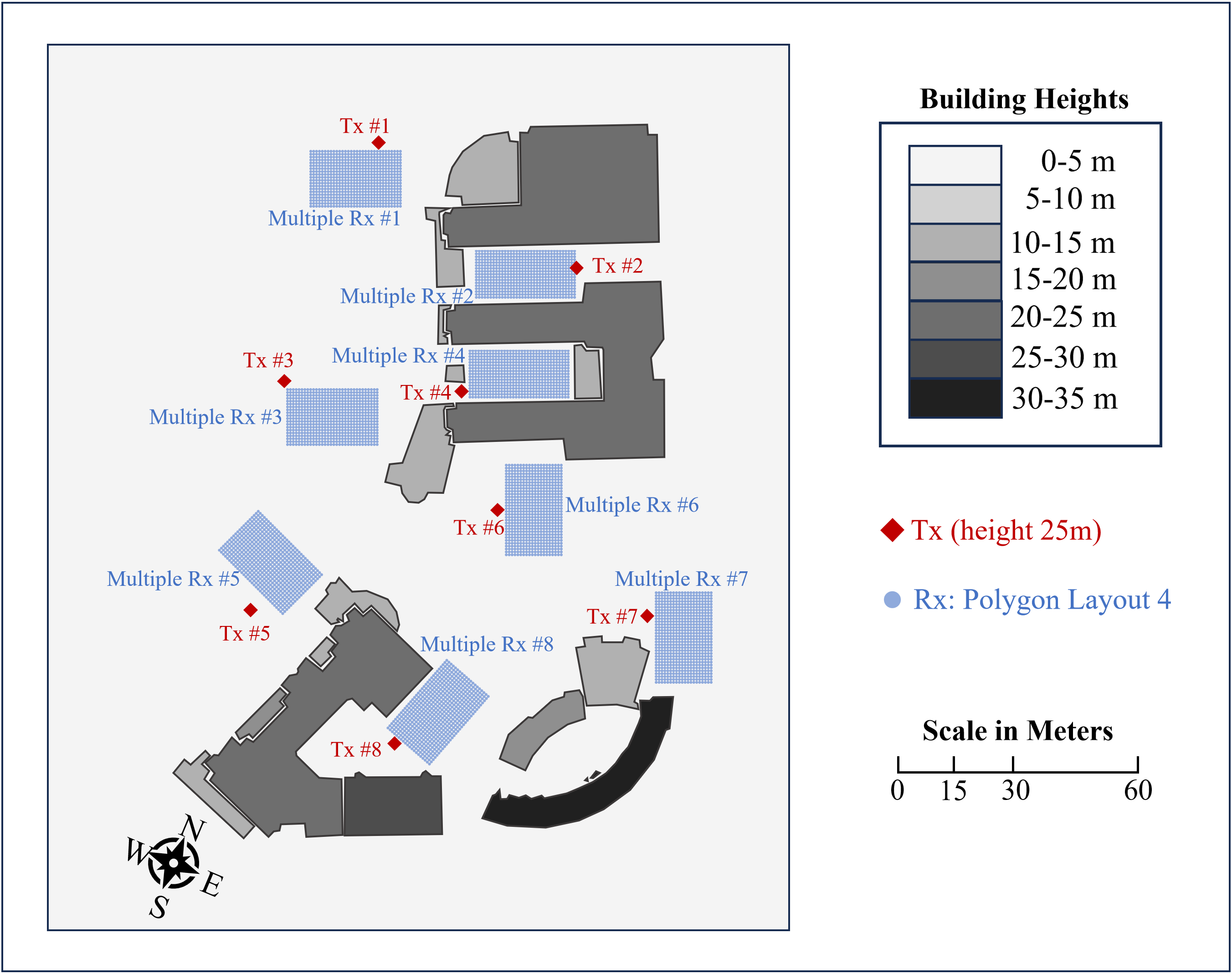}
        \caption{Bird’s eye view of Scenario 3 and the horizontal positions of Txs and Rxs.}
    \end{subfigure}

    \caption{Scenarios used to construct the dataset and the distribution of Txs and Rxs within these scenarios.}
    \label{fig:combined_scene_views}
\end{figure*}

We consider the following three real-world outdoor scenarios  for data generation:
\begin{itemize}
    \setlength{\itemsep}{0.05\baselineskip}
    \setlength{\parsep}{0pt}
    \setlength{\topsep}{0pt}
    \setlength{\partopsep}{0pt}
    \item \textbf{Scenario 1: Rosslyn.} This scenario, depicted in Figs.~\ref{fig:combined_scene_views}(a) and (b), represents a section of Rosslyn, Virginia, and is publicly available in \gls{WI} as an urban scenario. It occupies a region measuring \(150\,\text{m} \times 160\,\text{m} \times 80\,\text{m}\) (width × length × height), containing 10 buildings and 60 surfaces, each with sides perpendicular to the ground.
    
    \item \textbf{Scenario 2: SJTU-SEIEE.} This scenario, as depicted in Figs.~\ref{fig:combined_scene_views}(c) and \ref{fig:combined_scene_views}(d), is the School of Electronic Information and Electrical Engineering (SEIEE) on the Minhang campus of Shanghai Jiao Tong University (SJTU). It is reconstructed using an oblique photogrammetry-based monolithic modeling technique\footnote{This technology is primarily based on multi-angle aerial imagery to obtain high-precision mesh models, after which it employs deep learning and segmentation algorithms to automatically identify individual buildings or structures.}. It spans a \(195\text{ m} \times 255\text{ m} \times 30\text{ m}\) region and features a dragon-shaped layout consisting of 3 buildings and 1606 surfaces.
    
    \item \textbf{Scenario 3: SJTU-ME.} This scenario is the School of Mechanical Engineering on the Minhang campus of SJTU, as shown in Figs.~\ref{fig:combined_scene_views}(e) and \ref{fig:combined_scene_views}(f). This scenario is also reconstructed by the oblique photogrammetry-based monolithic modeling technique. It covers a region measuring \(195\,\text{m} \times 255\,\text{m} \times 30\,\text{m}\), comprising 6 buildings and 522 surfaces.
\end{itemize}
In these scenarios, each scatterer semantic class is associated with one of the 11 typical material types specified in the ITU-R standard \cite{Standard_ITU} for simplicity \footnote{Note that scatterer semantics contain more aspects than material type, but mainly material type is considered in this work as an initial trial.}, and the corresponding semantic classes are illustrated in different colors in Figs.~\ref{fig:combined_scene_views}(a), (c), and (e). Additionally, three humidity levels are considered, with the corresponding material parameters set as follows:
\begin{itemize}
    \setlength{\itemsep}{0.05\baselineskip}
    \setlength{\parsep}{0pt}
    \setlength{\topsep}{0pt}
    \setlength{\partopsep}{0pt}
    \item \textbf{Dry:} The conductivity and relative permittivity of all materials are set according to the specifications in \cite{Standard_ITU}, where the ground is configured as Very Dry Ground.
    \item \textbf{Medium Dry:} The ground is set as Medium Dry Ground, while the conductivity and relative permittivity of all other materials are increased by 30\% compared to the dry condition.
    \item \textbf{Wet:} The ground is set as Wet Ground, while the conductivity and relative permittivity of all other materials are increased by 60\% compared to the dry condition.
\end{itemize}

\subsubsection{Simulation Dataset Description}
\begin{figure}[!b]
    \centering
    \includegraphics[width=0.95\linewidth]{ 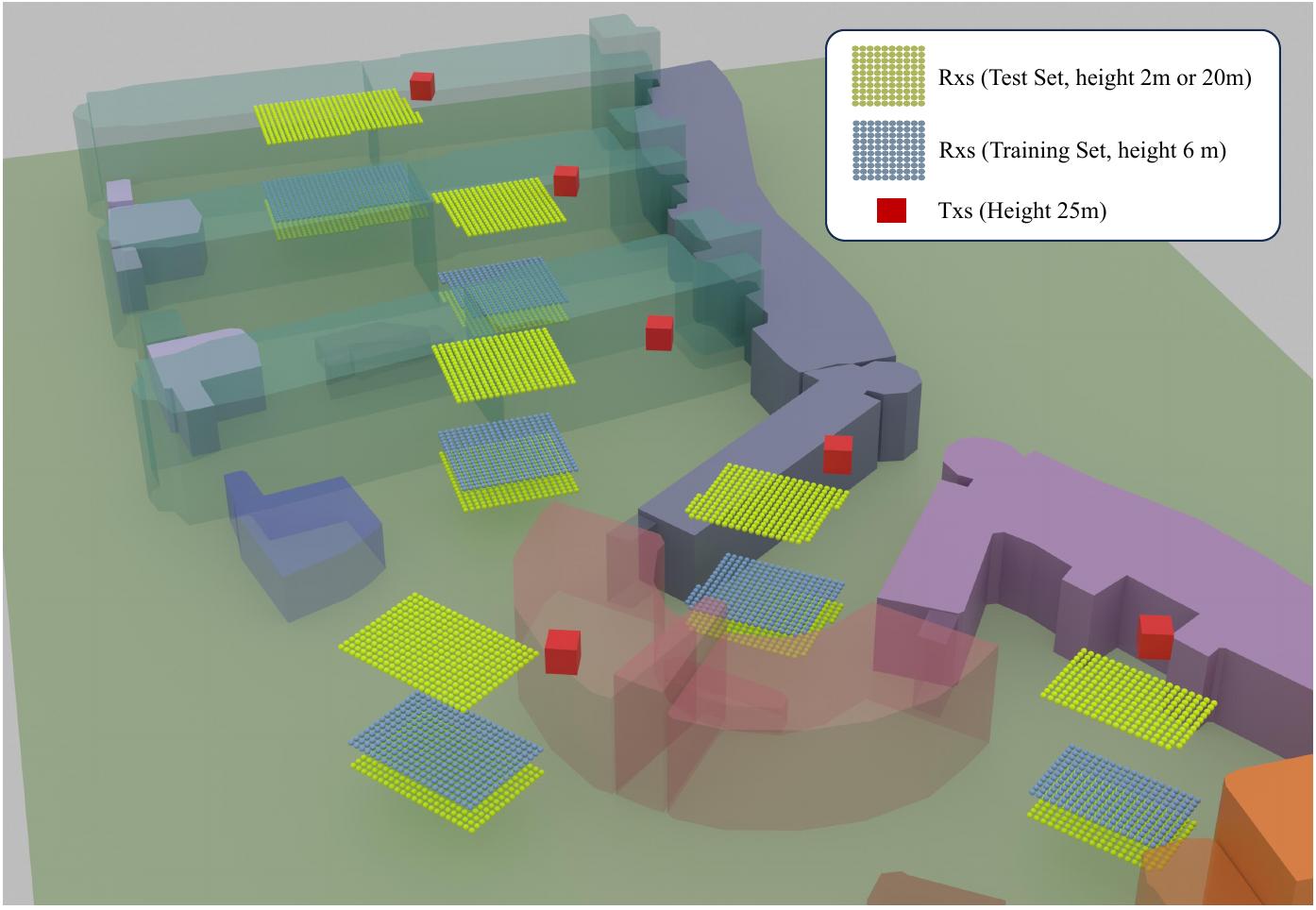}
    \caption{Distribution of Txs and Rxs in datasets \(\mathcal{S}_6\) to \(\mathcal{S}_{14}\). For clarity, only the lowest and highest Rxs are shown; the remaining Rxs are positioned at intermediate heights of 4\,m and from 8\,m to 20\,m in 2\,m increments. Certain buildings are intentionally blurred to emphasize Rxs' spatial locations.}
    \label{intra height}
\end{figure}

Based on the three scenarios described above, we construct three types of datasets to support different stages of training and evaluation: module-wise pre-training, system-wise end-to-end training, and generalization test. All datasets are generated under a consistent \gls{RT} configuration unless otherwise specified. In particular, rays are launched uniformly at $0.4^\circ$ intervals, and both the Tx and Rx are equipped with isotropic antenna\footnote{Note that the proposed approach is applicable to arbitrary antenna patterns by incorporating the corresponding antenna gain on each ray.} with a gain of 0 dBi. The transmission power is set to 0 dBm, and the carrier frequency is 3.5\,GHz. An overview of the constructed datasets is given in Table~\ref{WI Settings}, with construction details presented below:
\begin{itemize}
    \setlength{\itemsep}{0.05\baselineskip}
    \setlength{\parsep}{0pt}
    \setlength{\topsep}{0pt}
    \setlength{\partopsep}{0pt}
    \item \textbf{Module-wise pre-training:} To isolate specific polarization components, we construct $\mathcal{S}_\parallel$ and $\mathcal{S}_\perp$ from Scenario 1 using single-reflection paths. Txs and Rxs are placed at the same height (2\,m) and located near building edges. $\mathcal{S}_\parallel$ includes horizontally polarized signals reflected by buildings and vertically polarized signals reflected by the ground, while $\mathcal{S}_\perp$ consists of the opposite combinations.
    
    \item \textbf{System-wise end-to-end training:} The datasets $\mathcal{S}_1$-$\mathcal{S}_3$ constructed from Scenario 2 differ exclusively in the humidity levels. At this stage, we collect the CIR for each Tx–Rx pair as defined in \eqref{end-to-end feature}.
    
    \item \textbf{Generalization test:} To evaluate the model's generalization performance, we construct twelve challenging datasets covering two aspects: intra-scenario spatial transferability and inter-scenario zero-shot generalization capability. Datasets $\mathcal{S}_4$–$\mathcal{S}_{14}$ from Scenario 2 are constructed for the former. Specifically, $\mathcal{S}_4$ and $\mathcal{S}_5$ vary the horizontal positions of the Rxs while keeping their height fixed, whereas $\mathcal{S}_6$–$\mathcal{S}_{14}$ vary the Rx heights at fixed horizontal positions, as illustrated in Fig.~\ref{intra height}. Dataset $\mathcal{S}_{15}$ from Scenario 3 is constructed for the latter, where the original scenario is replaced with Scenario 3 to test cross-scenario generalization. Additionally, at this stage, we also collect the CIR for each Tx–Rx pair as defined in \eqref{end-to-end feature}.
        
\end{itemize}

\subsection{Simulation Setup}
\subsubsection{Network Architecture and Hyperparameters}
The outgoing ray prediction network introduced in Section III is detailed as follows. The output dimension of the embedding layer is set to 5, which is then expanded by a latent space MLP containing a single hidden layer with 32 neurons. The first FFN block consists of input and output layers with 64 neurons and a hidden layer with 32 neurons, while the second block comprises 32 neurons for both input and output, and a hidden layer with 16 neurons. The encoding dimension $L$ in the PosEnc is set to 4. ReLU is used as the nonlinear activation function throughout the network. To train this network, we adopt the Adam optimizer with a batch size of 64. For module-wise pre-training, the model is trained for 400 epochs with an initial learning rate of $1\text{e}^{-3}$, which is halved every 80 epochs. For system-wise end-to-end training, we use 200 training epochs, with an initial learning rate of $4\text{e}^{-4}$, reduced by half every 50 epochs. Weight parameters are initialized using the Xavier Uniform method. 


\vspace{-0.25\baselineskip}
\subsubsection{Evaluation Metrics}
After obtaining the ground truth (GT) label \( \mathcal{H}_n \) for each predicted \( \hat{\mathcal{H}}_n \) via the mapping defined in \eqref{Matching pre and label}, we assess the MPC prediction performance using the following three metrics:
\begin{itemize}
    \setlength{\itemsep}{0pt}
    \setlength{\parsep}{0pt}
    \setlength{\topsep}{0pt}
    \setlength{\partopsep}{0pt}
    \item \textbf{Overall prediction error (Overall Error):} This metric quantifies the overall modeling error across all propagation paths and channel elements. For each propagation path, we compute the sum of element-wise NMSEs between \( \hat{\mathcal{H}}_n \) and \( \mathcal{H}_n \), followed by averaging across all paths in all Tx–Rx pairs.
        
    \item \textbf{Reflection coefficient magnitude error (RCM Error):} This metric complements the overall error by specifically evaluating the prediction accuracy of ray-surface interaction loss. Both predicted and GT values are converted to a logarithmic scale, and the error is computed using \gls{NMSE}, following the same averaging procedure as in overall error.
        
    \item \textbf{Average delay error (AvgDelay Error):} This metric assesses the accuracy of the power delay profile (PDP) by comparing the predicted and GT average delays (AvgDelays) \cite{WiNeRT}. For each Tx–Rx pair, the AvgDelay is computed as the power-weighted mean of \( \tau_n \), with weights given by the linear-scale received powers \( p(a_n) \):
    \begingroup
    \setlength{\abovedisplayskip}{0.15\baselineskip}
    \setlength{\belowdisplayskip}{0.15\baselineskip}
    \setlength{\abovedisplayshortskip}{0.08\baselineskip}
    \setlength{\belowdisplayshortskip}{0.08\baselineskip}
    \begin{equation}
        \tau_{\text{avg}} = \frac{\sum_n p(a_n) \tau_n}{\sum_n p(a_n)}.
    \end{equation}
    \endgroup
    The \gls{MAE} between the predicted and GT values of \( \tau_{\text{avg}} \) is then computed and averaged over all Tx–Rx pairs.

\end{itemize}

\vspace{-0.3\baselineskip}
\subsubsection{Baselines} 
To evaluate GeNeRT's relative geometry, scatterer semantics, and polarization-aware modeling, we compare it with WiNeRT~\cite{WiNeRT} and LWDT~\cite{LWDT}, two neural \gls{RT} baselines. WiNeRT learns local ray-surface interactions in a differentiable ray-marching pipeline, whereas LWDT decouples geometric path tracing from \gls{EM} response learning and models object-level interactions. All methods use the same datasets, Tx--Rx settings, environment geometries, and matched-MPC metrics.

\begin{table*}[!tb]
    \centering
    \caption{Training-region MPC prediction performance of different approaches under varying humidity levels. Overall Error (Overall Err.) and RCM Error (RCM Err.) are in a logarithmic scale (dB), while AvgDelay Error (AvgDelay Err.) is in nanoseconds (ns).}
    \renewcommand{\arraystretch}{1.2}
    \setlength{\tabcolsep}{3.8pt}
    \footnotesize
    \begin{tabular*}{\textwidth}{@{\extracolsep{\fill}}lccccccccc@{}}
        \toprule
        & \multicolumn{3}{c}{$\mathcal{S}_1$ (Dry)} & \multicolumn{3}{c}{$\mathcal{S}_2$ (Medium Dry)} & \multicolumn{3}{c}{$\mathcal{S}_3$ (Wet)} \\
        \cmidrule(lr){2-4} \cmidrule(lr){5-7} \cmidrule(lr){8-10}
        & \makecell{Overall\\Err.} & \makecell{RCM\\Err.} & \makecell{AvgDelay\\Err.} & \makecell{Overall\\Err.} & \makecell{RCM\\Err.} & \makecell{AvgDelay\\Err.} & \makecell{Overall\\Err.} & \makecell{RCM\\Err.} & \makecell{AvgDelay\\Err.} \\
        \midrule
        WiNeRT & -34.55 & -22.43 & 5.84 &  -28.54 &  -17.93 & 10.58  & -28.95  & -18.61 & 9.49  \\
        LWDT & -38.04 & -26.32 & 3.82 & -38.13 & -24.93 & 4.25 & -37.46 & -23.66 & 5.10 
        \\
        \rowcolor{gray!15}
        GeNeRT & \textbf{-39.71} & \textbf{-27.07} & \textbf{3.43} & \textbf{-39.11} & \textbf{-26.68} & \textbf{3.56} & \textbf{-38.85} & \textbf{-26.62} & \textbf{3.85} \\
        \bottomrule
    \end{tabular*}
    \label{NMSE under Humidity Levels}
\end{table*}

\begin{figure*}[!t]
    \centering

    \begin{subfigure}[t]{0.32\linewidth}
        \centering
        \includegraphics[width=\linewidth]{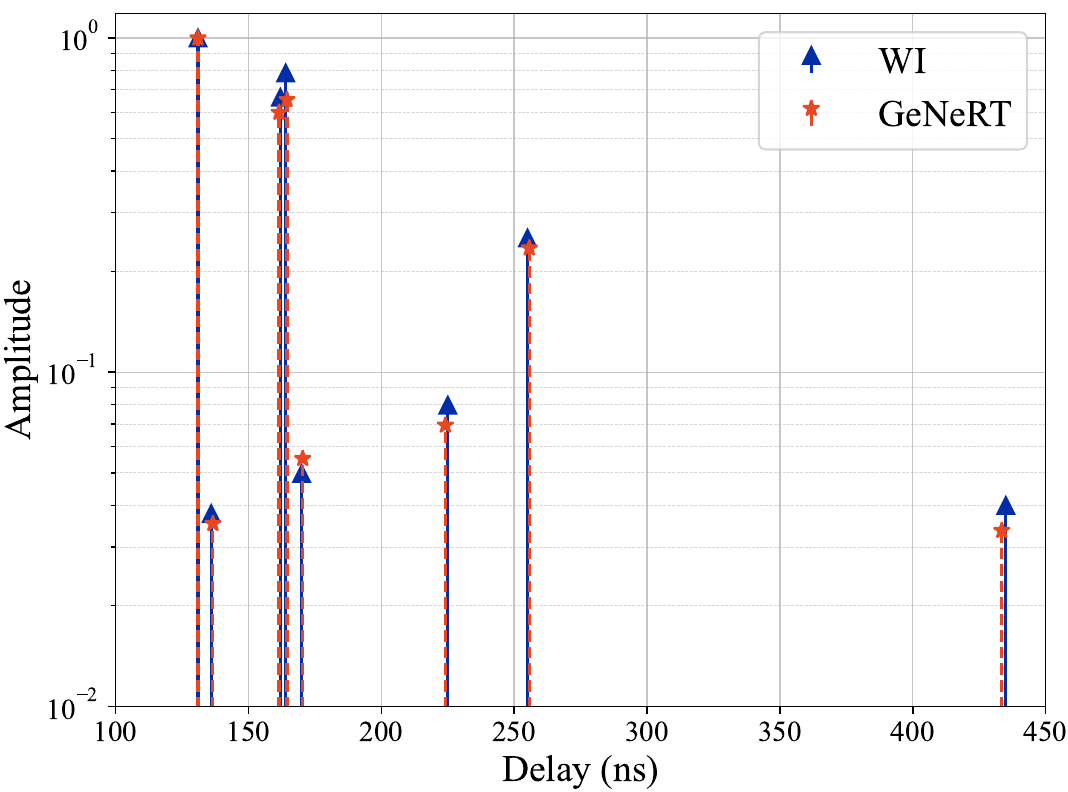}
        \caption{Tx--Rx pair \#1.}
    \end{subfigure}
    \begin{subfigure}[t]{0.32\linewidth}
        \centering
        \includegraphics[width=\linewidth]{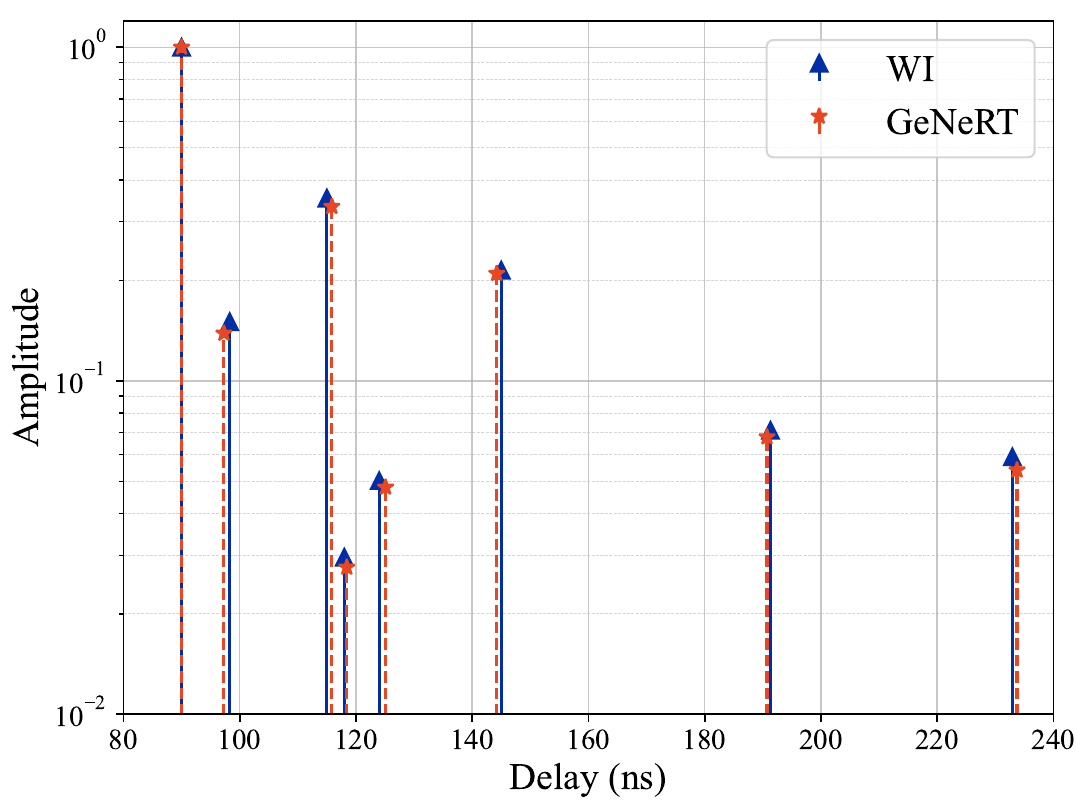}
        \caption{Tx--Rx pair \#2.}
    \end{subfigure}
    \begin{subfigure}[t]{0.32\linewidth}
        \centering
        \includegraphics[width=\linewidth]{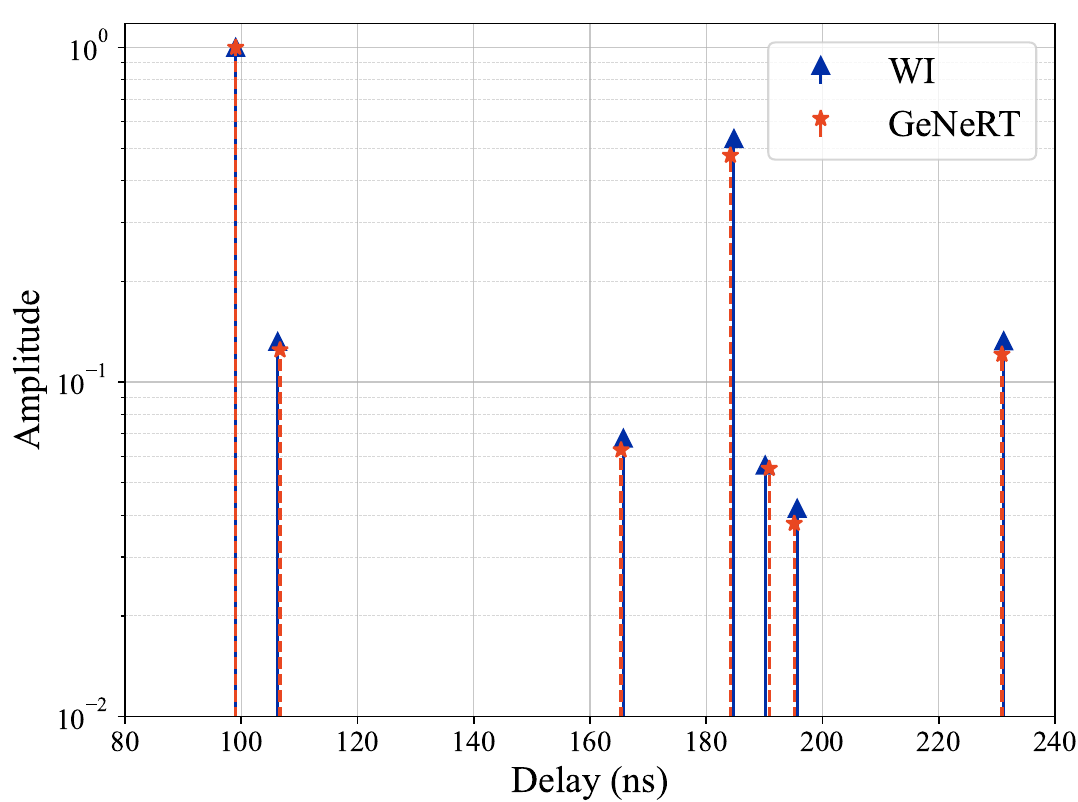}
        \caption{Tx--Rx pair \#3.}
    \end{subfigure}

    \caption{Comparison of CIRs between GeNeRT and WI under different Tx--Rx pairs.}
    \label{CIRs}
\end{figure*}

\vspace{-0.35\baselineskip}
\subsection{Training-Region MPC Prediction}\label{sec:training_region_results}

\begin{figure}[!t]
    \centering
    \includegraphics[width=0.95\linewidth]{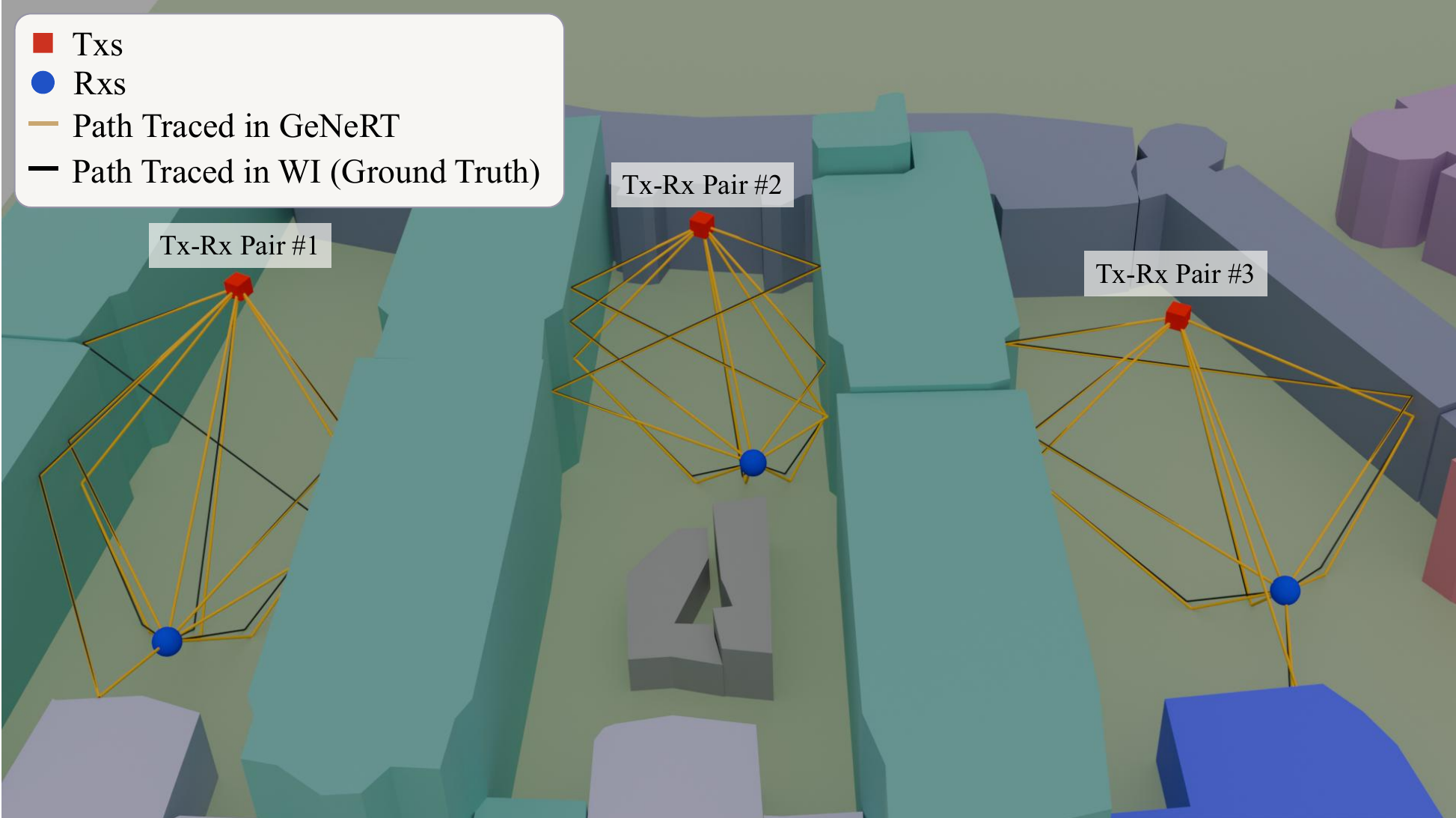}
    \caption{Propagation path diagram. Propagation paths produced by GeNeRT are generally consistent with those in WI, with certain paths exhibiting exact overlap.}
    \label{propogation_path_diagram}
    \vspace{-0.55\baselineskip}
\end{figure}

We evaluate the performance of GeNeRT in the test set of the $\mathcal{S}_1$ dataset. As shown in Table~\ref{NMSE under Humidity Levels}, GeNeRT outperforms both WiNeRT and LWDT across all evaluation metrics, demonstrating its superior accuracy in MPC prediction. It is worth noting that the overall error is significantly lower than the RCM error, mainly because it includes free-space propagation loss, which introduces minimal prediction error and dilutes the \gls{NMSE}. To provide a more intuitive illustration of GeNeRT's ability to predict MPCs, we take three Tx--Rx pairs from the $\mathcal{S}_1$ test set as examples and present their time-domain CIRs and propagation paths in Figs.~\ref{CIRs} and \ref{propogation_path_diagram}, respectively. As shown in Fig.~\ref{CIRs}, the delay values from WI and GeNeRT are nearly identical, except for several single-reflection paths, and the corresponding received power deviation remains within 1.5~dB. In Fig.~\ref{propogation_path_diagram}, the majority of paths predicted by GeNeRT closely align with those from WI, with only a few exhibiting minor deviations. These deviations consistently occur at the ray's final interaction with the environment, where the absence of the classical mirror-image method slightly perturbs the boundary condition enforcement and leads to sub-wavelength geometric discrepancies.

It is recognized that scatterers belonging to the same semantic class may exhibit diverse EM propagation behaviors under different humidity conditions. We further evaluate the adaptability of GeNeRT to varying humidity levels using the test sets of $\mathcal{S}_2$ and $\mathcal{S}_3$. As shown in Table~\ref{NMSE under Humidity Levels}, GeNeRT maintains stable MPC prediction across varying humidity levels. From the dry scenario $\mathcal{S}_1$ to the wet scenario $\mathcal{S}_3$, the variations in overall, RCM, and AvgDelay errors remain minimal, with respective standard deviations of only 0.43 dB, 0.23 dB, and 0.21 ns. In contrast, baseline methods suffer from noticeable degradation, particularly in AvgDelay error under high humidity. The robust performance of GeNeRT suggests that the system-wise end-to-end training effectively mitigates humidity-induced changes in scatterer characteristics, thereby ensuring reliable MPC prediction under diverse environmental conditions.

\subsection{Generalization Test}\label{sec:generalization_results}
\begin{table*}[!t]
    \centering
    \caption{Predictive performance of different approaches for generalization evaluation. Overall and RCM errors are in dB, while AvgDelay error is in ns.}
    \renewcommand{\arraystretch}{1.2}
    \setlength{\tabcolsep}{3.8pt}
    \footnotesize
    \begin{tabular*}{\textwidth}{@{\extracolsep{\fill}}lccccccccc@{}}
        \toprule
        & \multicolumn{3}{c}{$\mathcal{S}_4$} & \multicolumn{3}{c}{$\mathcal{S}_5$} & \multicolumn{3}{c}{$\mathcal{S}_{15}$} \\
        \cmidrule(lr){2-4} \cmidrule(lr){5-7} \cmidrule(lr){8-10}
        & \makecell{Overall\\Err.} & \makecell{RCM\\Err.} & \makecell{AvgDelay\\Err.} & \makecell{Overall\\Err.} & \makecell{RCM\\Err.} & \makecell{AvgDelay\\Err.} & \makecell{Overall\\Err.} & \makecell{RCM\\Err.} & \makecell{AvgDelay\\Err.} \\
        \midrule
        WiNeRT & -22.58 & -8.84 & 28.45 & -20.89 & -7.14 & 30.99 & -8.92 & -3.36 & 35.18 \\
        LWDT & -25.58 & -13.01 & 19.26 & -25.07 & -11.47 & 22.35 & -10.85 & -4.97 & 32.38 \\
        \rowcolor{gray!15}
        GeNeRT & \textbf{-37.69} & \textbf{-25.79} & \textbf{3.96} & \textbf{-36.29} & \textbf{-24.69} & \textbf{4.34} & \textbf{-35.36} & \textbf{-23.77} & \textbf{4.91} \\
        \bottomrule
    \end{tabular*}
    \label{Gen Test}
\end{table*}

\begin{figure}[!t]
    \centering
    \includegraphics[width=0.95\linewidth]{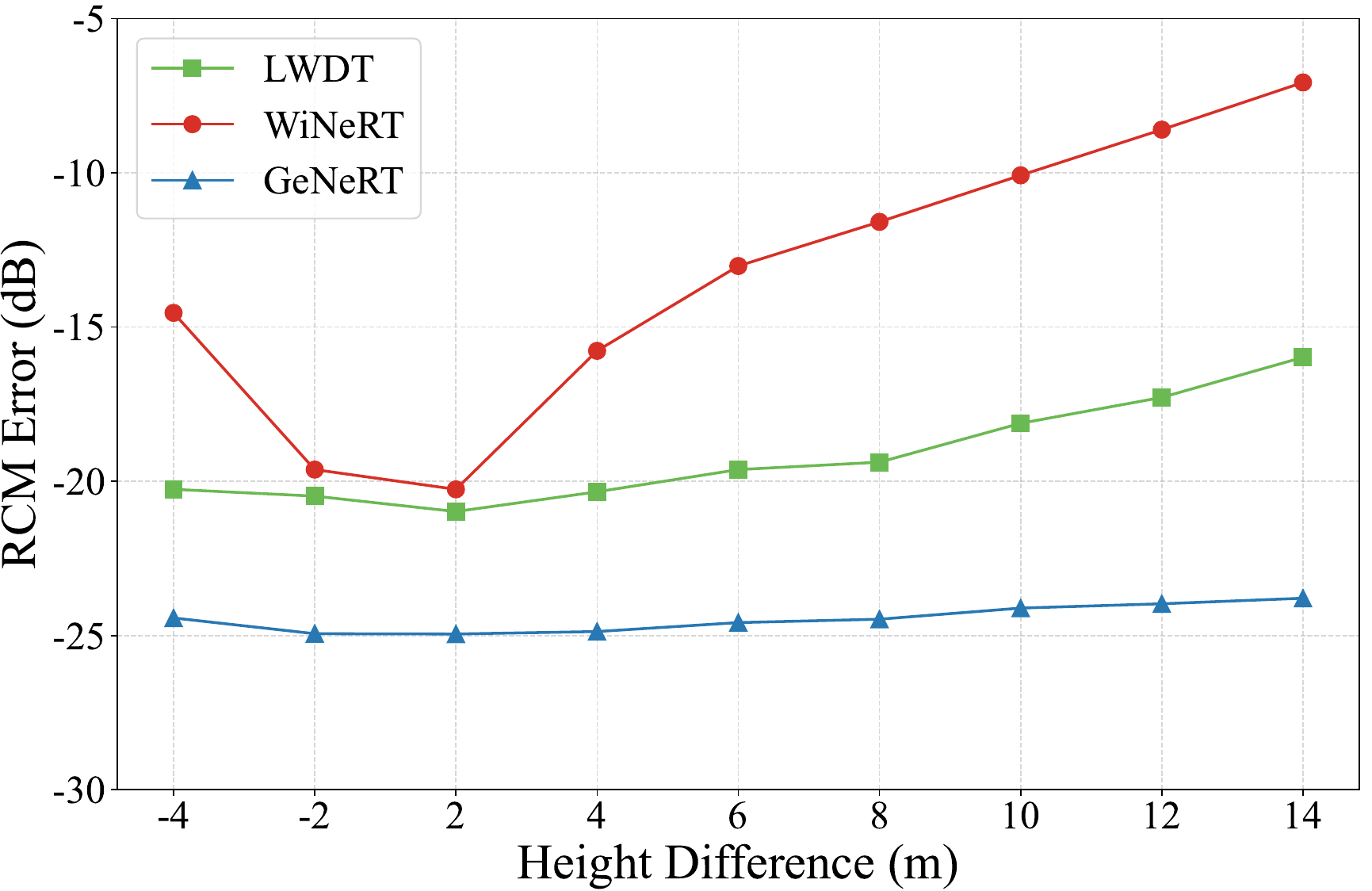}
    \caption{NMSE of RCM under different Rx heights. From left to right, the x-axis corresponds to datasets $\mathcal{S}_6$--$\mathcal{S}_{14}$ with increasing height difference, defined as the test Rx height minus the training Rx height. Positive and negative values indicate that the test Rx is higher and lower than the training Rx, respectively.}
    \label{NMSE Performance on Different Height}
    \vspace{-0.35\baselineskip}
\end{figure}

We evaluate the generalization capability of different approaches on datasets $\mathcal{S}_4$ to $\mathcal{S}_{15}$. Specifically, $\mathcal{S}_4$–$\mathcal{S}_{14}$ assess intra-scenario spatial transferability, while $\mathcal{S}_{15}$ tests inter-scenario zero-shot generalization.

\subsubsection{Intra-Scenario Spatial Transferability}

In system-wise end-to-end training, receivers are positioned within a limited region. This motivates an evaluation of intra-scenario spatial transferability to test the model’s generalization to untrained areas. To assess generalization along specific dimensions (e.g., horizontal and vertical) while avoiding confounding factors, we employ two controlled test settings. Datasets $\mathcal{S}_4$ and $\mathcal{S}_5$ vary the Rx’s horizontal position at a fixed height, and are used to evaluate horizontal generalization. In contrast, datasets $\mathcal{S}_6$ to $\mathcal{S}_{14}$ vary the Rx’s height at a fixed horizontal location, and are used to assess vertical generalization.

From the results of $\mathcal{S}_4$ and $\mathcal{S}_5$ in Table \ref{Gen Test}, GeNeRT consistently achieves the highest MPC prediction accuracy. When the Rx's horizontal position varies, GeNeRT exhibits only a slight drop across all metrics, while both baselines experience notable performance degradation compared to the training region. As shown in Fig. \ref{NMSE Performance on Different Height}, when the Rx height deviates from the training height, GeNeRT continues to deliver stable performance, maintaining an RCM error of -23.79 dB even at a height difference of 14 m. In contrast, LWDT maintains moderate accuracy at small height differences, but its error gradually increases as the Rx moves higher. WiNeRT is particularly sensitive to vertical changes, showing a sharp increase in error as the height difference grows. These results demonstrate the superior generalization capability of GeNeRT across both horizontal and vertical~spatial~variations.

\subsubsection{Inter-Scenario Zero-Shot Generalization}

After evaluating intra-scenario spatial transferability, we consider a more challenging case with entirely different building layouts. In campus scenarios, different building clusters often share common characteristics, such as materials and architectural styles, but their surface locations, orientations, and Tx--Rx visibility relationships can change substantially. A neural RT method that maintains reliable MPC prediction across such clusters can eliminate the need for retraining under scenario changes. As shown in the results of $\mathcal{S}_{15}$ in Table~\ref{Gen Test}, GeNeRT consistently maintains superior performance, achieving an overall error of $-35.36$ dB, an RCM error of $-23.77$ dB, and an AvgDelay error of only 4.91 ns.

The performance gap can be explained by the different ways in which the methods encode geometry and ray--surface interactions. WiNeRT relies more heavily on absolute spatial information and therefore tends to learn interaction patterns tied to the training layout; when the building layout changes, the distribution of ray directions, interaction locations, and surface normals shifts, leading to degraded path and attenuation prediction. LWDT improves transferability through relative geometric features and object-level modeling, but its object-specific representations are still learned from the training scenario and can capture only local interaction behavior under sparse outdoor training data. As a result, both baselines suffer from accumulated errors in the predicted interaction coefficients and propagation paths under the unseen scenario. In contrast, GeNeRT uses relative angular features, scatterer semantics, and polarization-aware interaction modeling, which allows the learned ray--surface interaction behavior to transfer across building clusters that share similar material and architectural characteristics. These results further validate the inter-scenario generalization capability of GeNeRT.

\subsection{Discussions on Runtime Efficiency and Ablation Study}

\subsubsection{Runtime Efficiency}
\setcounter{bottomnumber}{3}
\renewcommand{\bottomfraction}{0.95}
\renewcommand{\textfraction}{0.05}
\begin{figure}[!b]
    \centering
    \captionsetup{type=table}
    \captionof{table}{Tx/Rx count settings for different setups.}
    \label{tab:tx_rx_config}
    \setlength{\tabcolsep}{4.5pt}
    \footnotesize
    \begin{tabular}{c|c|c|c}
        \hline
        \textbf{Setup} & \textbf{Tx} & \textbf{Rx} & \textbf{Number of Tx--Rx pairs} \\
        \hline
        Setup 1 & Tx\#1 & Multiple Rx \#1--8 & 2352 \\
        Setup 2 & Tx\#1,2 & Multiple Rx \#1--4 & 2352 \\
        Setup 3 & Tx\#1--4 & Multiple Rx \#1,2 & 2352 \\
        Setup 4 & Tx\#1--8 & Multiple Rx \#1 & 2352 \\
        \hline
    \end{tabular}

    \vspace{0.85\baselineskip}
    \captionsetup{type=figure}
    \includegraphics[width=\linewidth,height=2.65in,keepaspectratio]{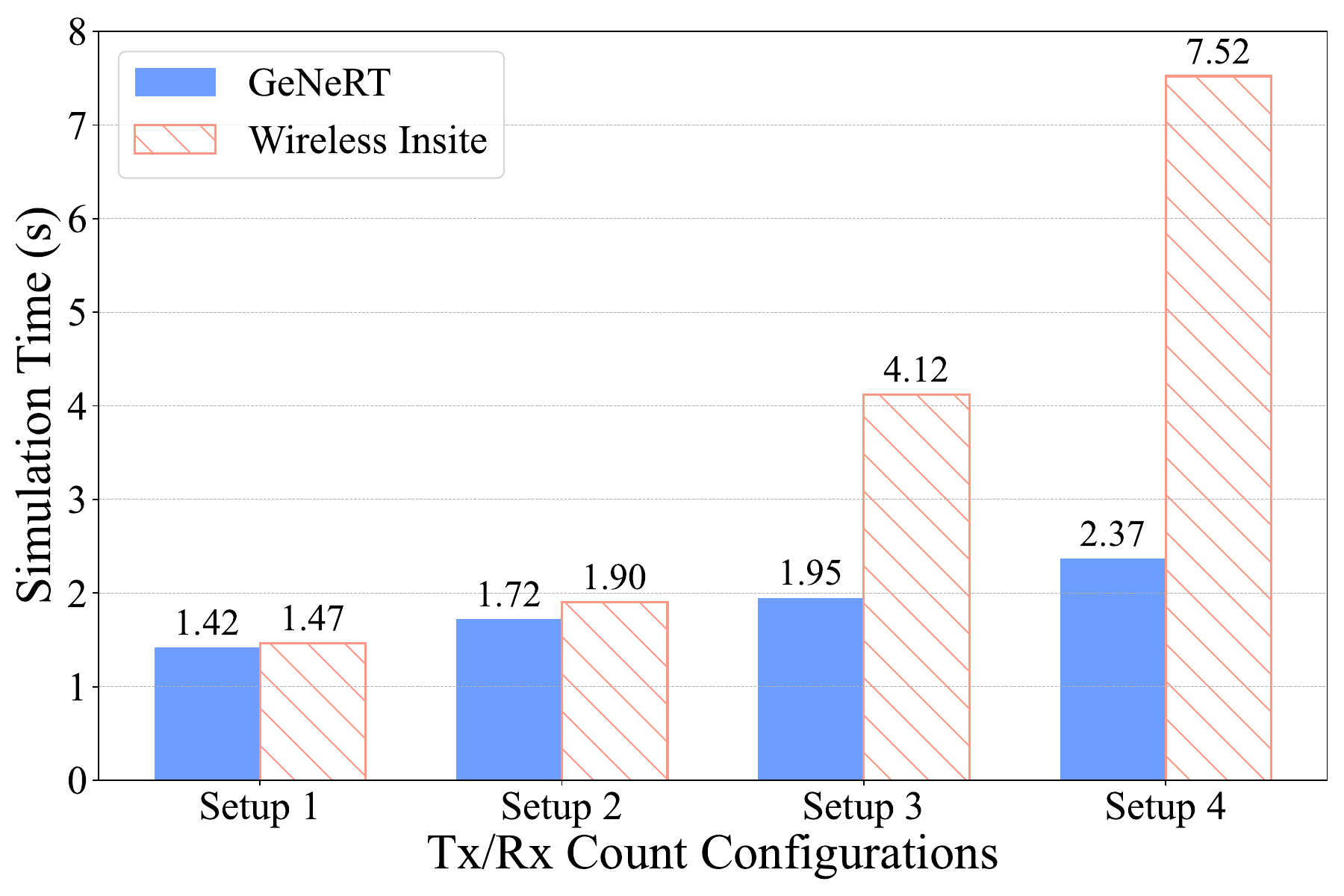}
    \captionof{figure}{Simulation time comparison between GeNeRT and WI.}
    \label{fig:time_overhead}
    \vspace{-0.55\baselineskip}
\end{figure}

We choose \gls{WI} as a representative of conventional RT methods for simulation time comparison. Specifically, we compare the time overhead of GeNeRT and WI under the $\mathcal{S}_{15}$ dataset configuration. We further examine how different numbers of Txs and Rxs affect simulation time by considering four Tx–Rx setups in Scenario 3. The specific settings are summarized in Table~\ref{tab:tx_rx_config}. While the total number of Tx–Rx pairs remains constant across all setups, the ratio of Txs to Rxs varies. For a fair comparison, simulation time is defined as the total duration required to compute the CIR for all Tx–Rx pairs. Both methods are executed on the same server equipped with an AMD EPYC 9654 CPU and an NVIDIA GeForce RTX 4090 GPU. Fig.~\ref{fig:time_overhead} presents the average simulation time over 100 independent runs under each setup. As observed, when the number of Txs is small (e.g., only one), the simulation time of WI and GeNeRT is comparable. However, as the proportion of Txs increases, GeNeRT exhibits a growing advantage, highlighting its superior runtime efficiency in large-scale multi-Tx scenarios.

\subsubsection{Ablation Study}

\begin{figure}[!b]
    \centering
    \includegraphics[width=\linewidth,height=3.00in,keepaspectratio]{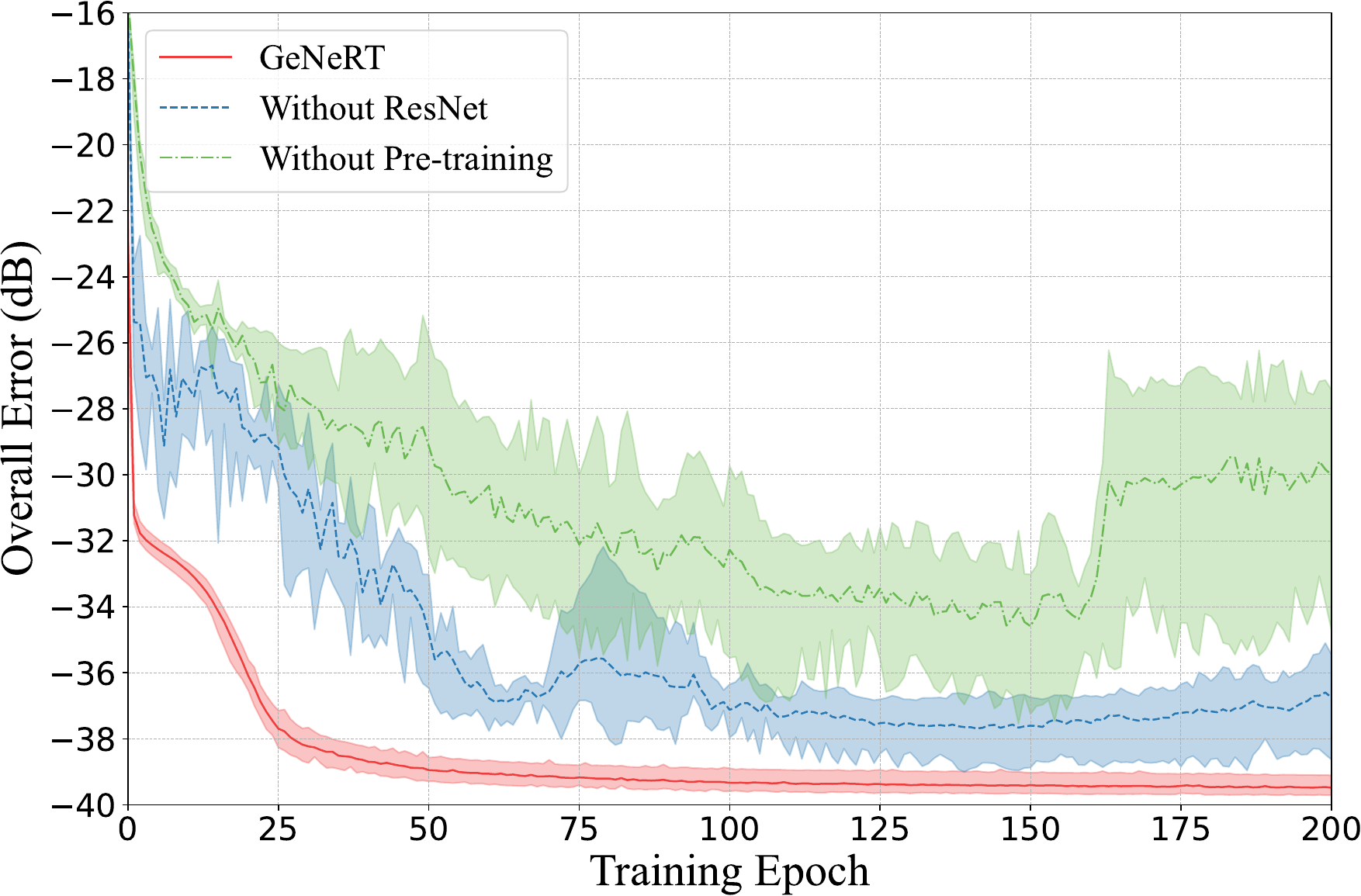}
    \caption{Ablation study of key network components.}
    \label{fig:ablation_experiment}
    \vspace{-0.9\baselineskip}
\end{figure}

We conduct ablation experiments on key components of the network to assess the effectiveness of the ResNet and the module-wise pre-training. Fig.~\ref{fig:ablation_experiment} presents the overall error of GeNeRT under three different network configurations, evaluated on the test set of $\mathcal{S}_1$ during system-wise end-to-end training with 10 different random seeds, where each configuration is represented by a low-opacity mean line and a high-opacity shaded error band. The three configurations are as follows:
\begin{itemize}
    \item \textbf{Without ResNet:} The skip connections in the network are removed, while the module-wise pre-training is retained. Only the polarization fusion module is active during training.
    \item \textbf{Without Pre-training:} The polarization component prediction module is trained from scratch, with its weights initialized using the Xavier Uniform method. Both the polarization fusion module and the polarization component prediction module are activated.
    \item \textbf{GeNeRT:} The full model proposed in this paper. Only the polarization fusion module is activated, with its weights initialized via module-wise pre-training.
\end{itemize}

As shown in Fig.~\ref{fig:ablation_experiment}, GeNeRT consistently achieves the lowest error throughout training, demonstrating the effectiveness of both the ResNet architecture and the module-wise pre-training strategy. Removing either component leads to clear performance degradation. In particular, the absence of pre-training causes significantly higher error and unstable convergence, especially after 160 epochs. The absence of ResNet also results in a noticeable performance drop compared to the full GeNeRT model, though it remains more stable than the model trained without pre-training.

\section{Measurement-Based Fine-Tuning}

{\looseness=1
The simulation results in Section IV demonstrate that GeNeRT can accurately predict MPCs under different simulation settings and generalize to unseen regions and scenarios. Nevertheless, practical propagation environments inevitably introduce real-world features that are difficult to fully capture in simulation, such as material uncertainty, surface roughness, and construction details. To further validate the practical applicability of GeNeRT, this section presents a measurement-based fine-tuning experiment conducted in a real outdoor campus scenario. The objective is to examine whether a small number of measured MPCs can further adapt the simulation-trained GeNeRT model to real-world propagation characteristics.\par}

\begin{figure*}[!b]
    \centering
    \begin{subfigure}[c]{0.48\textwidth}
        \vspace{0pt}
        \centering
        \includegraphics[width=\linewidth]{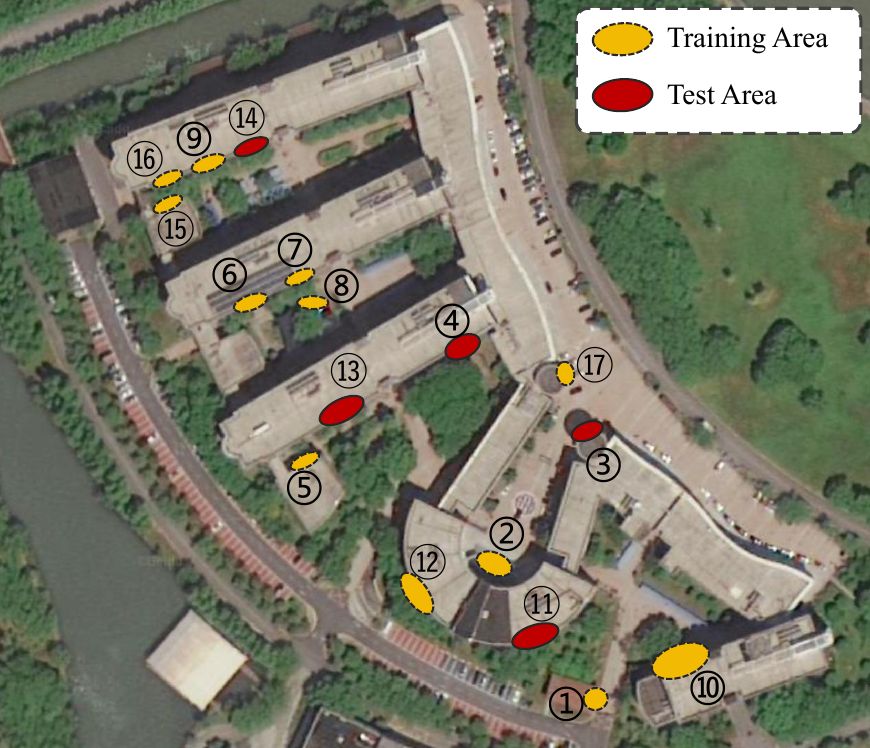}
        \caption{Measurement locations in the SJTU-SEIEE environment. The orange and red elliptical regions denote the fine-tuning and test measurement areas, respectively. Each elliptical region contains multiple Tx--Rx pairs, whose detailed deployment patterns are shown in (b) and (c).}
        \label{fig:measurement_area}
    \end{subfigure}
    \hfill
    \begin{minipage}[c]{0.48\textwidth}
        \vspace{0pt}
        \centering
        \begin{subfigure}[t]{\linewidth}
            \vspace{0pt}
            \centering
            \includegraphics[width=\linewidth]{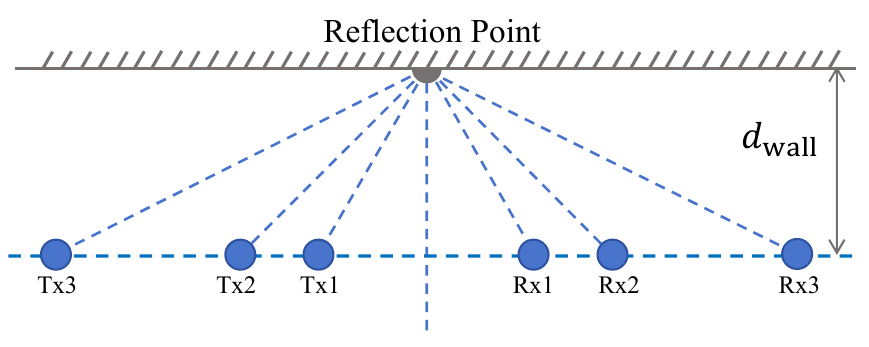}
            \caption{Tx--Rx deployment for locations 1--9. Three configurations are measured at each location, where Tx$_i$ and Rx$_i$ form the measured pair for the $i$-th configuration. These configurations correspond to wall-reflection incident angles of $30^\circ$, $45^\circ$, and $60^\circ$ and extract the LoS, ground-reflected, and wall-reflected MPCs.}
            \label{fig:measurement_layout_a}
        \end{subfigure}
        \vspace{1.5mm}
        \begin{subfigure}[t]{\linewidth}
            \vspace{0pt}
            \centering
            \includegraphics[width=\linewidth]{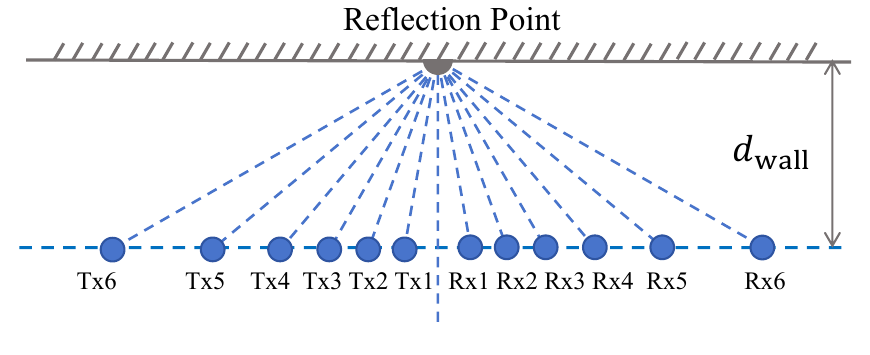}
            \caption{Tx--Rx deployment for locations 10--17. Six configurations are measured at each location, where Tx$_i$ and Rx$_i$ form the measured pair for the $i$-th configuration. These configurations cover wall-reflection incident angles of $10^\circ$--$60^\circ$ with a $10^\circ$ interval and extract the LoS and wall-reflected MPCs.}
            \label{fig:measurement_layout_b}
        \end{subfigure}
    \end{minipage}

    \caption{Measurement campaign design. (a) Measurement locations and fine-tuning/test regions in the SJTU-SEIEE environment. (b) and (c) Tx--Rx deployment patterns for the two measurement-location groups, where same-index Tx/Rx nodes, e.g., Tx1--Rx1 and Tx2--Rx2, are paired during measurement. The values of $d_{\mathrm{wall}}$ for all locations are listed in Table~\ref{tab:measurement_dwall}.}
    \label{fig:measurement_campaign}

    \vspace{1.5mm}

    {\captionsetup{type=table}
    \caption{Wall-distance settings for the measurement locations. The parameter $d_{\mathrm{wall}}$ denotes the perpendicular distance from the Tx/Rx placement line to the wall.}
    \label{tab:measurement_dwall}
    \renewcommand{\arraystretch}{1.15}
    \setlength{\tabcolsep}{3.3pt}
    \footnotesize
    \begin{tabular*}{\textwidth}{@{\extracolsep{\fill}}c*{17}{c}@{}}
        \toprule
        Location index & 1 & 2 & 3 & 4 & 5 & 6 & 7 & 8 & 9 & 10 & 11 & 12 & 13 & 14 & 15 & 16 & 17 \\
        \midrule
        $d_{\mathrm{wall}}$ (m) & 3 & 3 & 3 & 3 & 3 & 3 & 3 & 3 & 4 & 3 & 5 & 3 & 3 & 3.5 & 2.5 & 3.5 & 3 \\
        \bottomrule
    \end{tabular*}}
\end{figure*}

\begingroup
\tightdbltopfloat
\begin{figure*}[!t]
    \centering
    \begin{minipage}[t]{0.35\linewidth}
        \centering
        \includegraphics[width=0.95\linewidth]{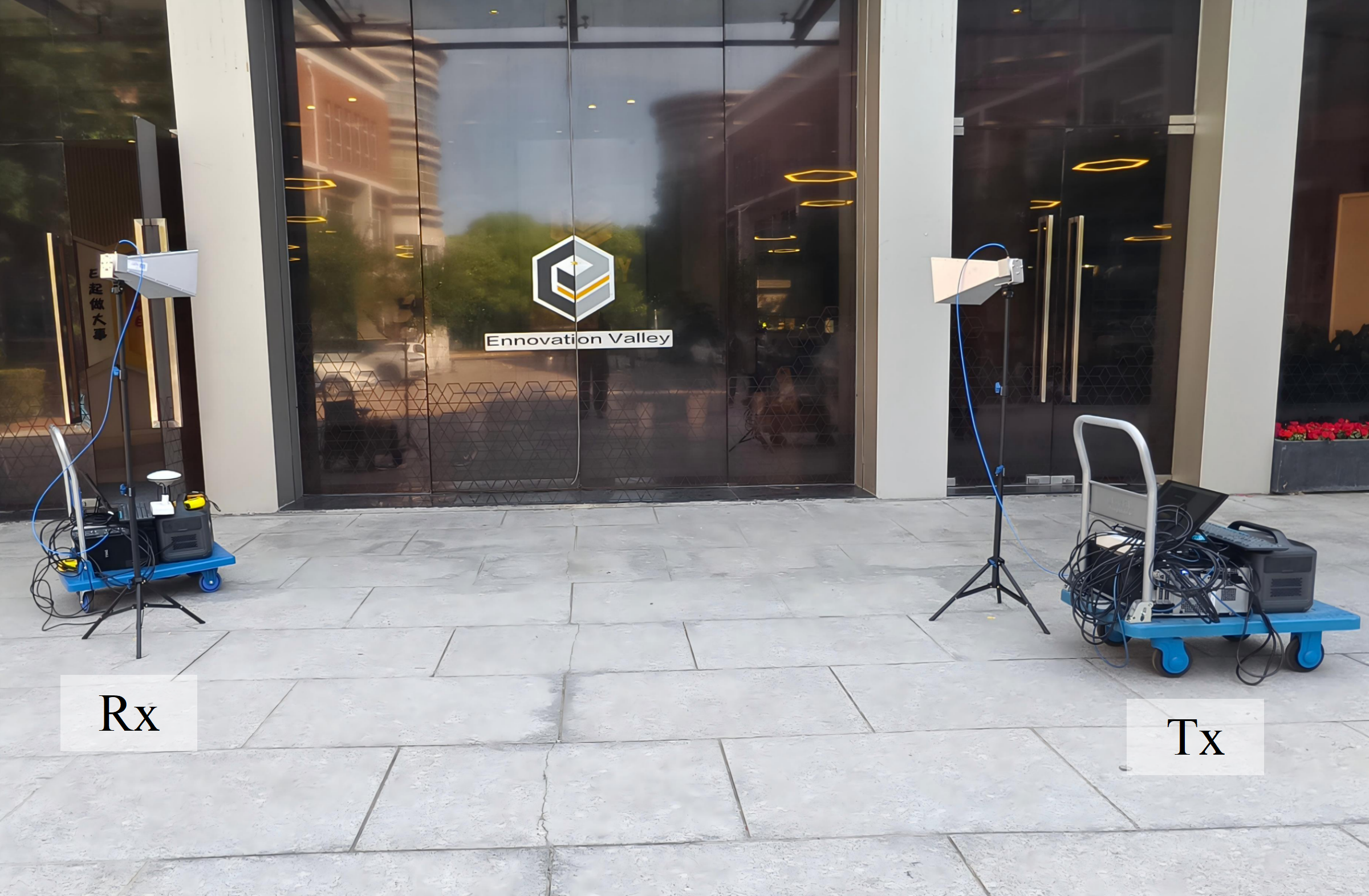}
        \vspace{0.5mm}
        \centerline{(a) Representative Tx--Rx deployment}
    \end{minipage}
    \hspace{0.10\linewidth}
    \begin{minipage}[t]{0.35\linewidth}
        \centering
        \includegraphics[width=0.85\linewidth]{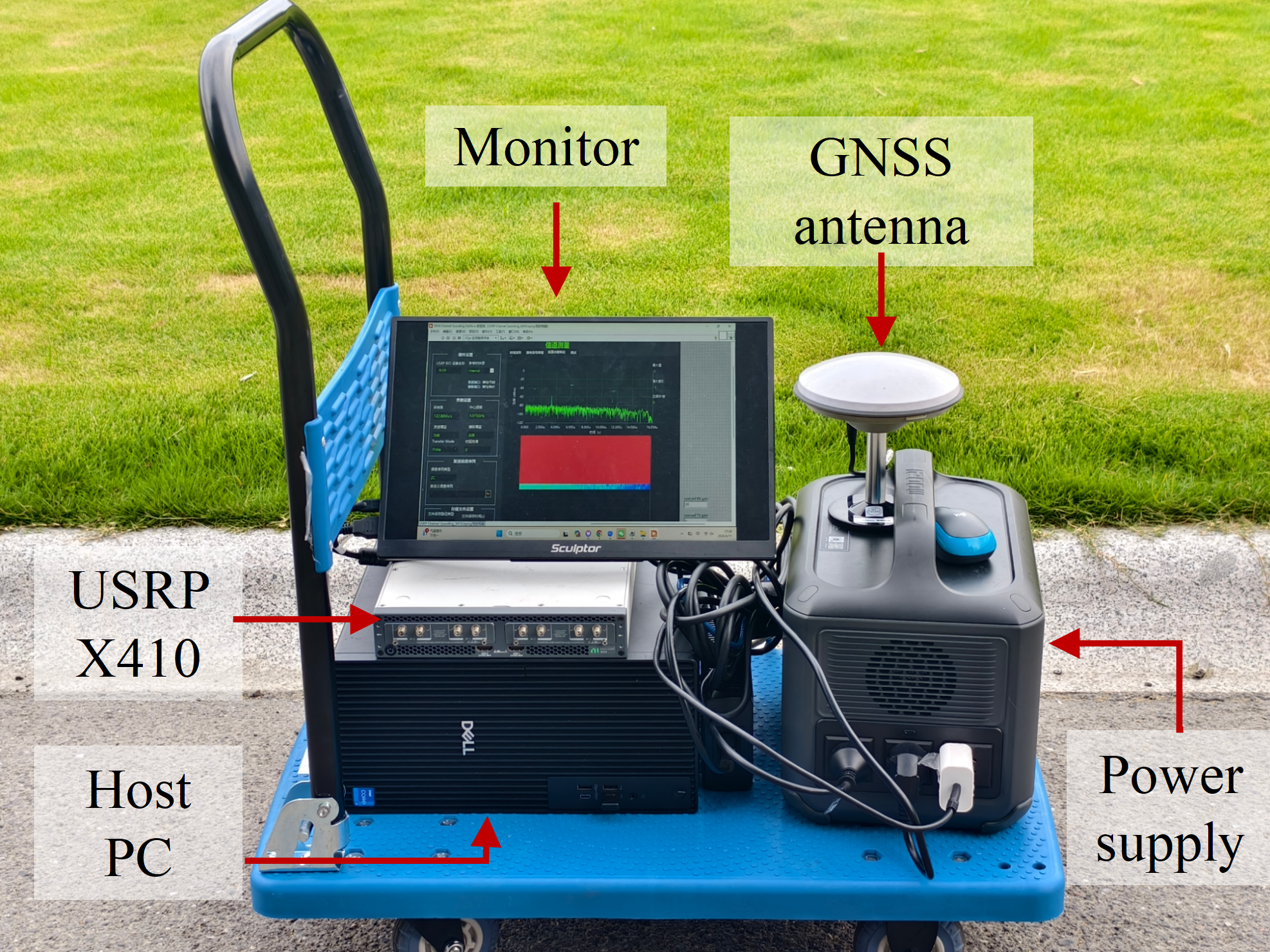}
        \vspace{0.5mm}
        \centerline{(b) Hardware architecture of the Rx}
    \end{minipage}
    \caption{Measurement hardware setup. (a) Representative long-range view of the measurement setup used throughout the measurement campaign, shown at Position~2 in Fig.~\ref{fig:measurement_campaign}(a), with the Tx and Rx located on the right and left sides, respectively. (b) Detailed modules of the Rx equipment. The Tx equipment is similar to that of the Rx.}
    \label{fig:measurement_equipment}
    \vspace{-0.6\baselineskip}
\end{figure*}

\subsection{Measurement Campaign}

The measurement campaign was conducted around the SEIEE at SJTU, i.e., the same physical environment corresponding to the SJTU-SEIEE digital twin shown in Figs.~\ref{fig:combined_scene_views}(c) and (d). As illustrated in Fig.~\ref{fig:measurement_campaign}(\subref{fig:measurement_area}), measurement data were collected at 17 different locations distributed across the SEIEE buildings.

In total, 102 reflected MPCs are extracted from the measurement campaign. The measurement locations are divided into two groups according to the Tx--Rx deployment pattern based on the available measurement space. For locations 1--9, three Tx--Rx configurations are measured at each location. For each configuration, the line-of-sight (LoS) component, the ground-reflected component, and the wall-reflected component are extracted. The corresponding incident angles of the wall-reflected paths are set to $30^\circ$, $45^\circ$, and $60^\circ$, respectively, as shown in Fig.~\ref{fig:measurement_campaign}(\subref{fig:measurement_layout_a}). For locations 10--17, six Tx--Rx configurations are measured at each location. For each configuration, the LoS component and the wall-reflected component are extracted, where the incident angles of the wall-reflected paths were set to $10^\circ$, $20^\circ$, $30^\circ$, $40^\circ$, $50^\circ$, and $60^\circ$, respectively, as shown in Fig.~\ref{fig:measurement_campaign}(\subref{fig:measurement_layout_b}). Here, $d_{\mathrm{wall}}$ denotes the perpendicular distance from the Tx/Rx placement line to the wall, and its location-dependent values are summarized in Table~\ref{tab:measurement_dwall}. This measurement design provides reflected MPC samples under different propagation distances, reflection surfaces, and incident angles, thereby enabling the fine-tuning process to capture practical ray-surface interaction characteristics.

The measurement setup is shown in Fig.~\ref{fig:measurement_equipment}. The channel sounder consisted of two USRP X410 software-defined radios, with one configured as the Tx and the other as the Rx. The channel measurements were conducted at a carrier frequency of 3.5 GHz, with a transmit power of 12 dBm. Vertically polarized directional antennas with a gain of 12.5 dBi were employed at both the Tx and Rx sides. During the measurements, the Tx and Rx antennas were mounted at a height of 1.5\,m above the ground and placed according to the designed geometric configurations. The antennas were directed at their respective reflection points. For wall reflections, this was the point illustrated in Figures \ref{fig:measurement_campaign}(b) and \ref{fig:measurement_campaign}(c), while for ground reflections, the antennas were aimed at the ground reflection point. The deployment geometry was designed to ensure that all reflection points lay in the far-field regions of both the transmitting and receiving antennas. The received MPCs were extracted from the measured PDPs. The measured LoS paths provide reference components for each Tx--Rx pair, while the measured reflected paths are used to evaluate and fine-tune the ray-surface interaction behavior of GeNeRT.

\subsection{Fine-Tuning Setup}

After collecting the measured MPCs, we split them according to the measurement regions shown in Fig.~\ref{fig:measurement_campaign}(\subref{fig:measurement_area}). The MPCs collected in the orange regions are used as the fine-tuning set, containing 75 measured reflected MPCs. The MPCs collected in the red regions are used as the test set, containing 27 measured reflected MPCs.

The fine-tuning procedure follows the measured-data fine-tuning strategy described in Section III-C. Starting from the simulation-trained GeNeRT model, we keep the geometry-related modules frozen and update only the modules that directly determine the electromagnetic interaction coefficients, namely the polarization component prediction module and the polarization fusion module. This design preserves the geometric propagation prior learned from simulation while adapting the ray-surface interaction parameters to real-world measurements.

For each measured Tx--Rx pair, the simulation-trained GeNeRT model first generates the predicted MPC set. Since the predicted and measured MPCs are unordered, the matching strategy in \eqref{Matching pre and label} is used to associate the predicted MPCs with the measured MPCs based on delay and angular consistency. The matched measured MPCs are then used as labels for fine-tuning. During this process, the learning rates of the polarization fusion module and the polarization component prediction module are set to $5\times10^{-4}$ and $5\times10^{-5}$, respectively. The model is trained for 60 epochs.

\FloatBarrier
\subsection{Fine-Tuning Results}

To evaluate the effect of measurement-based fine-tuning, we compare the simulation-trained GeNeRT model with the measurement-fine-tuned model on the measured test set. The quantitative MPC prediction errors are summarized in Table~\ref{tab:measurement_finetuning}, and representative PDP comparisons are shown in Fig.~\ref{fig:measurement_pdp}.

\begin{figure*}[!t]
    \centering
    \begin{minipage}[t]{0.32\textwidth}
        \centering
        \includegraphics[width=\linewidth]{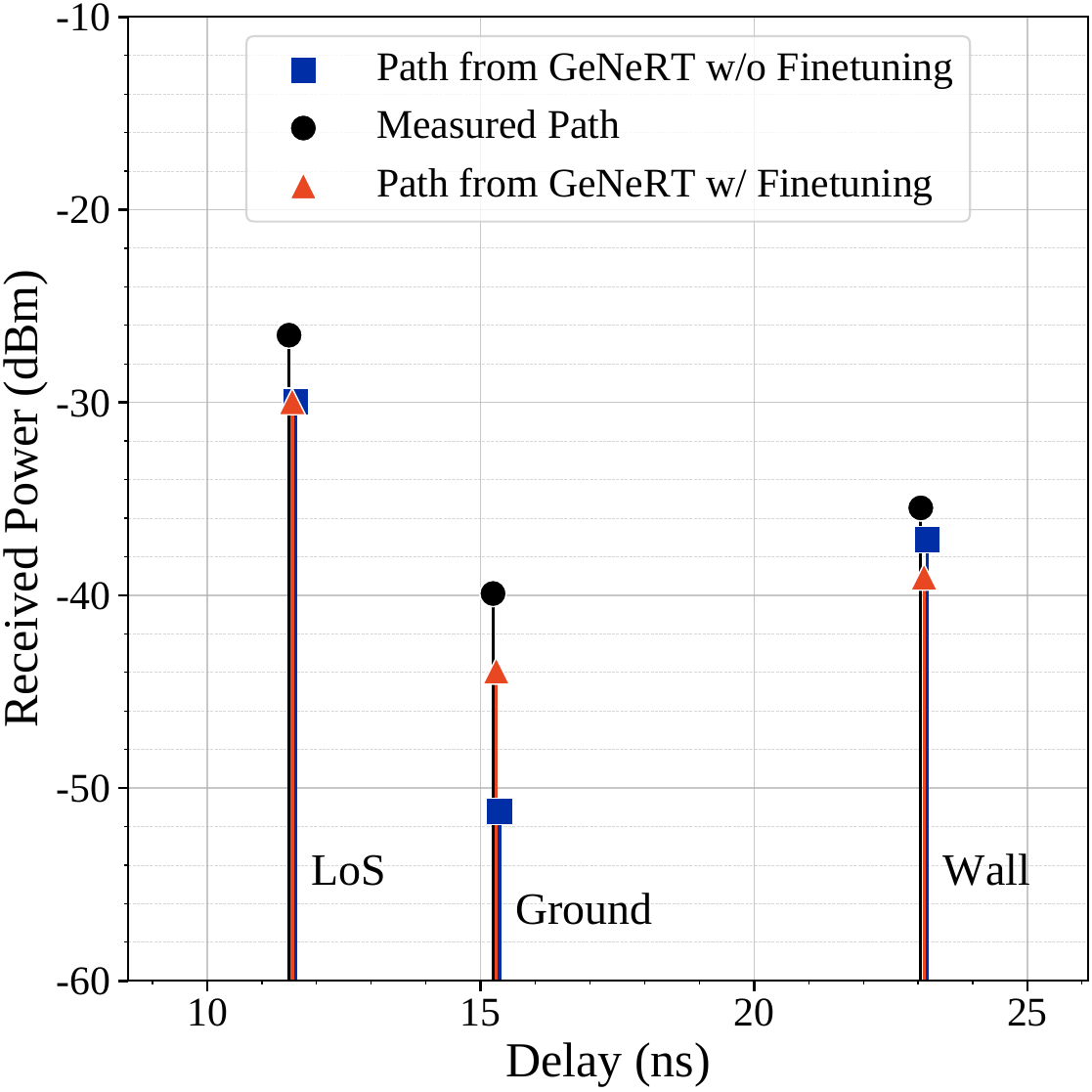}
        \vspace{1mm}
        \centerline{(a) Tx--Rx pair \#1}
    \end{minipage}
    \hfill
    \begin{minipage}[t]{0.32\textwidth}
        \centering
        \includegraphics[width=\linewidth]{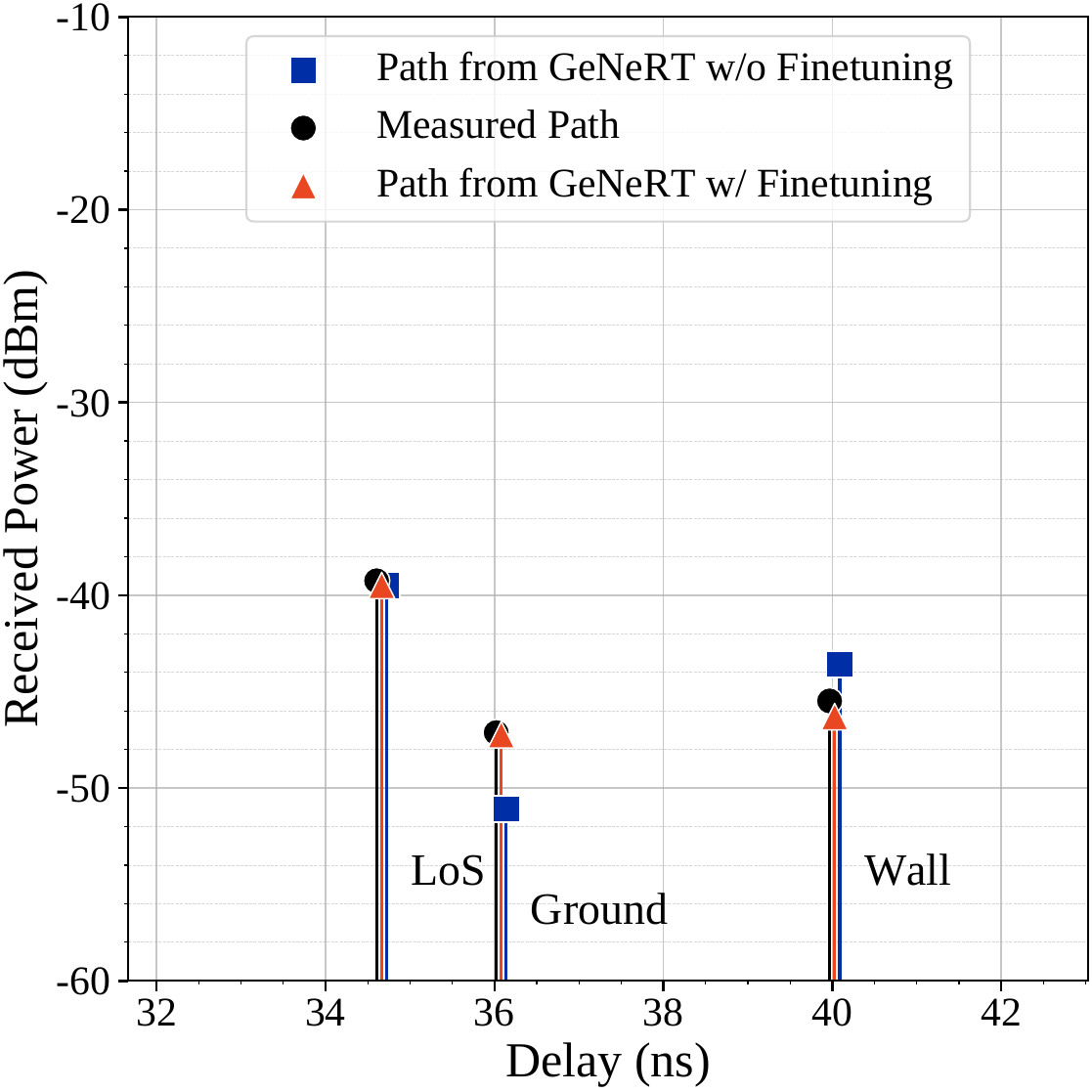}
        \vspace{1mm}
        \centerline{(b) Tx--Rx pair \#2}
    \end{minipage}
    \hfill
    \begin{minipage}[t]{0.32\textwidth}
        \centering
        \includegraphics[width=\linewidth]{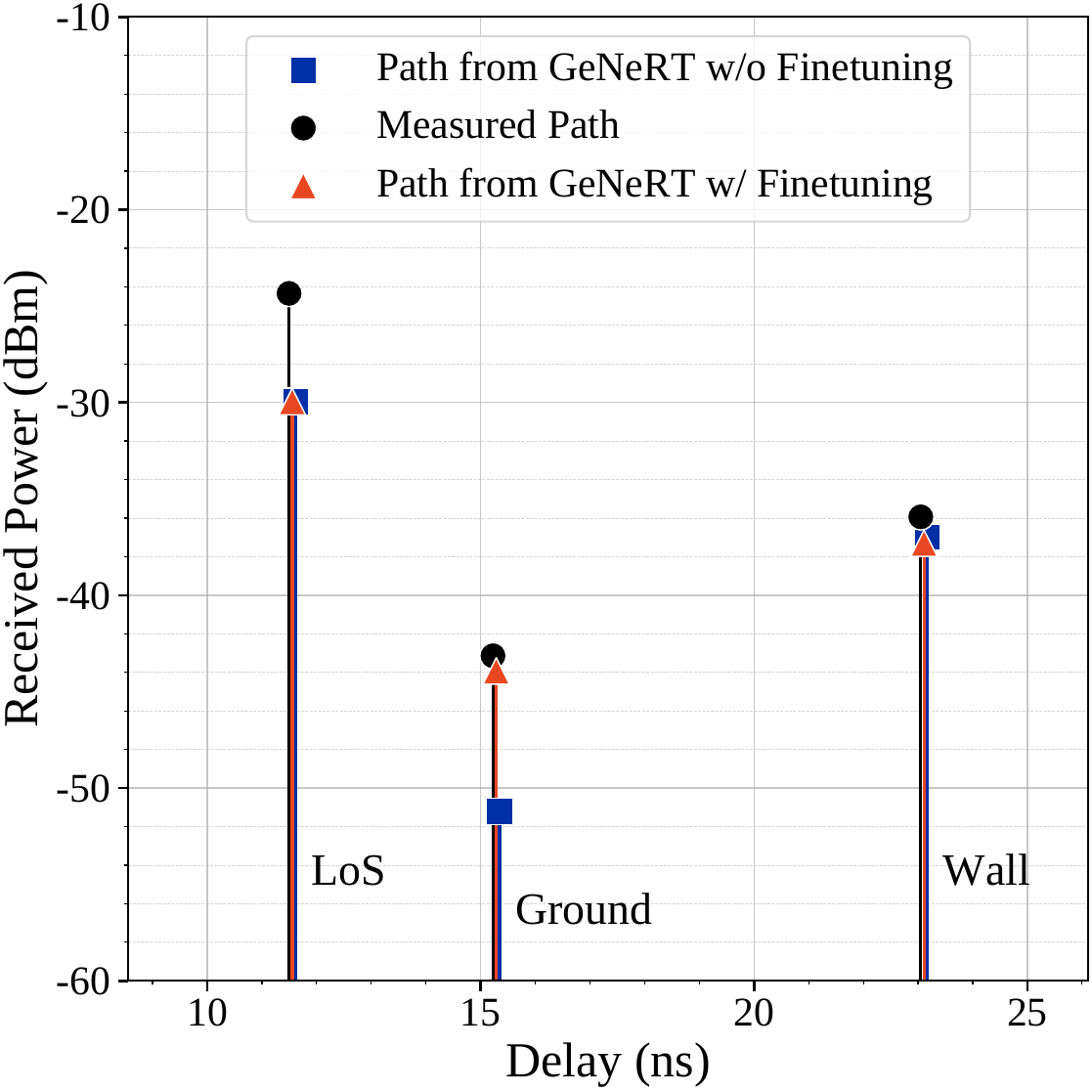}
        \vspace{1mm}
        \centerline{(c) Tx--Rx pair \#3}
    \end{minipage}
    \caption{Representative PDP comparisons among measured data, the simulation-trained GeNeRT model, and the measurement-fine-tuned GeNeRT model.}
    \label{fig:measurement_pdp}
    \vspace{-0.7\baselineskip}
\end{figure*}
\endgroup

\begin{table}[!t]
    \centering
    \caption{MPC prediction performance on the measured test set before and after measurement-based fine-tuning. Overall Error and RCM Error are in dB, while AvgDelay Error is in ns.}
    \label{tab:measurement_finetuning}
    \begin{tabular}{c|ccc}
        \hline
        Method & Overall Error & RCM Error & AvgDelay Error \\
        \hline
        \makecell{w/o Measurement\\Finetuning} & $-14.48$ & $-0.90$ & $6.28$ \\
        \hline
        \makecell{w/ Measurement\\Finetuning}  & $-22.90$ & $-9.32$ & $3.58$ \\
        \hline
    \end{tabular}
\end{table}

As shown in Table~\ref{tab:measurement_finetuning}, measurement-based fine-tuning substantially improves all evaluation metrics. Specifically, the RCM error decreases from $-0.90$ dB to $-9.32$ dB, indicating that the fine-tuned model can more accurately characterize the reflection loss of real scatterer surfaces. The overall error is also improved from $-14.48$ dB to $-22.90$ dB. In addition, the AvgDelay error is reduced from $6.28$ ns to $3.58$ ns. Since the AvgDelay metric is computed from the power-weighted delay distribution, this improvement indicates that the fine-tuned model not only predicts reflection magnitudes more accurately but also generates PDPs that are more consistent with real measurements.

Fig.~\ref{fig:measurement_pdp} further presents representative PDP comparisons for three measured Tx--Rx configurations. The measured PDPs are compared with the PDPs generated by the simulation-trained model and the measurement-fine-tuned model. Before fine-tuning, the predicted paths can follow the main propagation structure, but noticeable power deviations remain for several reflected components. After fine-tuning, the predicted PDPs become much closer to the measured ones, especially for the wall-reflected and ground-reflected components. This demonstrates that the measured fine-tuning step effectively compensates for the mismatch between simulation-assumed and real-world electromagnetic interaction characteristics.

The above results verify the effectiveness of measured data in adapting GeNeRT to practical propagation environments. With only 75 measured reflected MPCs for fine-tuning, GeNeRT achieves clear performance gains in the untrained test regions. This indicates that the proposed physics-informed architecture can retain the general propagation knowledge learned from simulation, while the measurement-based fine-tuning step efficiently calibrates the electromagnetic interaction parameters to real-world measurements. Therefore, GeNeRT provides a practical and data-efficient solution for bridging simulation-based neural RT and real-world wireless channel modeling.

\section{Conclusion}

The results show that GeNeRT can improve neural RT by combining a deterministic propagation pipeline with a physics-informed learnable interaction module. The use of relative geometric features and scatterer semantics reduces spatial dependence, while the Fresnel-inspired dual-branch design introduces an electromagnetic prior for polarization-dependent ray-surface interactions. This combination explains why GeNeRT maintains strong MPC prediction accuracy in the training region, remains stable across humidity settings, and transfers to both untrained receiver regions and a different outdoor scenario.

The measurement-based fine-tuning experiment further indicates that simulation-trained neural RT can be adapted to practical propagation environments with sparse measured MPCs. By updating only the polarization-related interaction modules, the model preserves the geometric propagation prior learned from simulation while recalibrating the electromagnetic response of real scatterers. This data-efficient adaptation is useful for outdoor wireless digital twins, where complete measured path geometry is difficult to obtain and where material properties, surface roughness, construction details, and measurement-system imperfections may deviate from the simulation assumptions.

Several limitations remain. The present study focuses on reflection as the foundational propagation mechanism, so future work should extend the neural interaction prediction module to diffuse scattering, diffraction, and mixed propagation mechanisms. The current scatterer semantics mainly encode material type, which is only a first approximation of the electromagnetic behavior of real surfaces. More fine-grained semantic modeling could integrate images, geometric attributes, material priors, and measured channel data to distinguish scatterers with similar labels but different propagation responses. Expanding the measurement campaign across more sites, frequencies, antenna patterns, and environmental conditions would further clarify the operating range of GeNeRT.

\bibliographystyle{IEEEtran}
\bibliography{GeNeRT_v11}

@inproceedings{conferenceWork,
  author    = {Kejia Bian and others},
  title     = {Generalizable Neural Ray Tracing Towards Physics-Informed Intelligent Channel Modeling},
  booktitle = {Proc. IEEE/CIC Int. Conf. Commun. China (ICCC)},
  year      = {2025},
  volume={},
  number={},
  pages={1-6}}

@misc{remcom_wireless_insite,
  author       = {{Remcom, Inc.}},
  title        = {{Wireless InSite}: {3D} Wireless Prediction Software},
  howpublished = {Remcom, Inc., State College, PA, USA. [Online]. Available: \url{https://www.remcom.com/wireless-insite-propagation-software}}
  }

@article{li2025deeprt,
  title={{DeepRT}: A Hybrid Framework Combining Large Model Architectures and Ray Tracing Principles for {6G} Digital Twin Channels},
  author={Li, Mingyue and others},
  journal={Electronics},
  volume={14},
  number={9},
  pages={1849},
  month=may,
  year={2025},
  publisher={MDPI}
}

@ARTICLE{9713743,
  author={Huang, Chen and others},
  journal={IEEE Trans. Antennas Propag.}, 
  title={Artificial Intelligence Enabled Radio Propagation for Communications—{Part II}: Scenario Identification and Channel Modeling}, 
  month=Feb,
  year={2022},
  volume={70},
  number={6},
  pages={3955-3969}}

@ARTICLE{10930391,
  author={Bakirtzis and others},
  journal={IEEE Wireless Commun.}, 
  title={Empowering Wireless Network Applications with Deep Learning-Based Radio Propagation Models}, 
  month=Mar,
  year={2025},
  volume={},
  number={},
  pages={1-8}}

@ARTICLE{10599118,
  author={Zhu, Ethan and others},
  journal={IEEE Wireless Commun.}, 
  title={Physics-Informed Generalizable Wireless Channel Modeling with Segmentation and Deep Learning: Fundamentals, Methodologies, and Challenges}, 
  year={2024},
  volume={31},
  number={6},
  pages={170-177}}

@article{Eertmans2026TransformInvariant,
  author  = {J{\'e}rome Eertmans and Enrico M. Vitucci and Vittorio Degli-Esposti and Nicola Di Cicco and Laurent Jacques and Claude Oestges},
  title   = {Transform-Invariant Generative Ray Path Sampling for Efficient Radio Propagation Modeling},
  journal = {arXiv preprint arXiv:2603.01655},
  year    = {2026}
}

@INPROCEEDINGS{mildenhall2020nerf,
  author={Mildenhall, Ben and Srinivasan, Pratul P. and Tancik, Matthew and Barron, Jonathan T. and Ramamoorthi, Ravi and Ng, Ren},
  booktitle={Proc. Eur. Conf. Comput. Vis. (ECCV)},
  title={{NeRF}: Representing Scenes as Neural Radiance Fields for View Synthesis},
  year={2020},
  volume={},
  number={},
  pages={405-421},
  doi={10.1007/978-3-030-58452-8_24}}

@INPROCEEDINGS{tancik2020fourier,
  author={Tancik, Matthew and Srinivasan, Pratul P. and Mildenhall, Ben and Fridovich-Keil, Sara and Raghavan, Nithin and Singhal, Utkarsh and Ramamoorthi, Ravi and Barron, Jonathan T. and Ng, Ren},
  booktitle={Proc. Adv. Neural Inf. Process. Syst. (NeurIPS)},
  title={Fourier Features Let Networks Learn High Frequency Functions in Low Dimensional Domains},
  year={2020},
  volume={33},
  number={},
  pages={7537-7547}}

@INPROCEEDINGS{jin2024sandwich,
  author={Jin, Yifei and others},
  booktitle={Proc. IEEE Int. Conf. Mach. Learn. Commun. Netw. (ICMLCN)}, 
  title={{SANDWICH}: Towards an Offline, Differentiable, Fully-Trainable Wireless Neural Ray-Tracing Surrogate}, 
  year={2025},
  volume={},
  number={},
  pages={1-7}}

@ARTICLE{10633853,
  author={Graglia, Roberto D. and Peterson, Andrew F.},
  journal={IEEE Antennas Propag. Mag.}, 
  title={Computational Electromagnetics and the {IEEE} Antennas and Propagation Society: Seventy-five years of shared history, part 1}, 
  month=Aug,
  year={2024},
  volume={66},
  number={5},
  pages={16-30}}

@ARTICLE{7508965,
  author={Steinböck, Gerhard and others},
  journal={IEEE Trans. Antennas Propag.}, 
  title={Hybrid Model for Reverberant Indoor Radio Channels Using Rays and Graphs}, 
  month=Jul,
  year={2016},
  volume={64},
  number={9},
  pages={4036-4048}}

@ARTICLE{9722715,
  author={Gupta, Ankit and others},
  journal={IEEE Trans. Antennas Propag.}, 
  title={Machine Learning-Based Urban Canyon Path Loss Prediction Using 28 {GHz} Manhattan Measurements}, 
  month=Feb,
  year={2022},
  volume={70},
  number={6},
  pages={4096-4111}}

@ARTICLE{9354041,
  author={Levie, Ron and others},
  journal={IEEE Trans. Wireless Commun.}, 
  title={{RadioUNet}: Fast Radio Map Estimation With Convolutional Neural Networks}, 
  month=Feb,
  year={2021},
  volume={20},
  number={6},
  pages={4001-4015}}

@INPROCEEDINGS{7063558,
  author={Maltsev, Alexander and others},
  booktitle={Proc. IEEE Globecom Workshops (GC Wkshps)}, 
  title={Quasi-deterministic approach to {mmWave} channel modeling in a non-stationary environment}, 
  year={2014},
  month = dec,
  volume={},
  number={},
  pages={966-971}}

@ARTICLE{10949588,
  author={Bai, Lu and others},
  journal={IEEE Commun. Surveys Tuts.}, 
  title={Multi-Modal Intelligent Channel Modeling: A New Modeling Paradigm via Synesthesia of Machines}, 
  year={2025},
  volume={},
  number={}}

@inproceedings{WRF_GS,
  author    = {C. Wen and others},
  title     = {{WRF-GS}: Wireless radiation field reconstruction with 3{D} gaussian splatting},
  booktitle = {Proc. IEEE INFOCOM},
  month=Feb,
  year      = {2025},
  pages     = {1--10}
}

@inproceedings{zhao2023nerf2,
  author    = {X. Zhao and others},
  title     = {Ne{RF}$^2$: Neural radio-frequency radiance fields},
  booktitle = {Proc. ACM MOBICOM},
  month=oct,
  year      = {2023},
  pages     = {1--15},
  address   = {Madrid, Spain}
}

@inproceedings{lu2024newrf,
  author    = {H. Lu and others},
  title     = {Ne{WRF}: A deep learning framework for wireless radiation field reconstruction and channel prediction},
  booktitle = {Int. Conf. on Machine Learning (ICML)},
  month=jun,
  year      = {2024},
  pages     = {1--13},
  address   = {Vienna, Austria}
}

@techreport{3gpp38901,
  author       = "{3GPP}",
  title        = "{Technical specification group radio access network; study on channel model for frequencies from 0.5 to 100 GHz (Release 14)}",
  institution  = "{3rd Generation Partnership Project (3GPP)}",
  type         = "TR",
  number       = "38.901 V14.2.0",
  year         = "2017",
  month        = "Sept.",
  url          = "http://www.3gpp.org/DynaReport/38901.htm"
}

@ARTICLE{omnidirectional_CIR,
  author={Samimi, Mathew K. and Rappaport, Theodore S.},
  journal={IEEE Trans. Microw. Theory and Techn.}, 
  title={3-{D} Millimeter-Wave Statistical Channel Model for 5{G} Wireless System Design}, 
  year={Jul. 2016},
  volume={64},
  number={7},
  pages={2207-2225}}

@article{moller2005fast,
  author    = {T. Möller and B. Trumbore},
  title     = {Fast, minimum storage ray-triangle intersection},
  journal   = {J. Graph. Tools},
  volume    = {2},
  number    = {1},
  pages     = {21--28},
  month=Sep,
  year      = {1997}
}

@INPROCEEDINGS{Sionna,
  author={Hoydis, Jakob and others},
  booktitle={Proc. IEEE Globecom Workshops (GC Wkshps)}, 
  title={Sionna {RT}: Differentiable Ray Tracing for Radio Propagation Modeling}, 
  month=Dec,
  year={2023},
  volume={},
  number={},
  pages={317-321}}

@book{rappaport2024wireless,
  author    = {T. S. Rappaport},
  title     = {Wireless Communications: Principles and Practice},
  edition   = {2nd},
  year      = {2001},
  publisher = {Prentice-Hall},
  address   = {Upper Saddle River, NJ, USA}
}

@inproceedings{ResNet,
  title={Deep residual learning for image recognition},
  author={He, Kaiming and others},
  booktitle={Proc. IEEE Conf. Comput. Vis. Pattern Recognit. (CVPR)},
  pages={770--778},
  year={Jun. 2016}
}

@misc{Standard_ITU,
  author    = {{P. Series}},
  title     = {Effects of Building Materials and Structures on Radiowave Propagation above about 100 {MHz}},
  year      = {2015},
  howpublished = {Recommendation ITU-R P.2040-1},
  pages     = {2040-1}
}

@article{wang2018survey,
  title={A survey of 5{G} channel measurements and models},
  author={Wang, Cheng-Xiang and others},
  journal={IEEE Commun. Surveys Tuts.},
  volume={20},
  number={4},
  pages={3142--3168},
    month  = aug,
  year={2018},
  publisher={IEEE}
}

@ARTICLE{he2018design,
  author={He, Danping and others},
  journal={IEEE Commun. Surveys Tuts.}, 
  title={The Design and Applications of High-Performance Ray-Tracing Simulation Platform for 5{G} and Beyond Wireless Communications: A Tutorial}, 
  month  = aug,
  year={2019},
  volume={21},
  number={1},
  pages={10-27}}

@inproceedings{WiNeRT,
  title={{WiNeRT}: Towards neural ray tracing for wireless channel modelling and differentiable simulations},
  author={Orekondy, Tribhuvanesh and others},
  booktitle={Proc. 11th Int. Conf. Learn. Rep.},
  pages = {1--20},
  month=feb,
  year={2023}
}

@article{LWDT,
  title={Learnable wireless digital twins: Reconstructing electromagnetic field with neural representations},
  author={Jiang, Shuaifeng and others},
  journal={IEEE Open J. Commun. Soc.},
  year={2025},
  publisher={IEEE}
}

@INPROCEEDINGS{Pi_Yibo,
  author={Jia, Haifeng and others},
  booktitle={Proc. IEEE Global Commun. Conf. (GLOBECOM)}, 
  title={Neural Reflectance Fields for Radio-Frequency Ray Tracing},
  month=Dec,
  year={2024},
  volume={},
  number={},
  pages={4226-4231}}

\end{document}